\documentclass{ahs}

\usepackage{cite}
\usepackage{latexsym}
\usepackage{dblfloatfix}    
\usepackage{lipsum}
\usepackage{wrapfig}
\usepackage{amsmath}
\usepackage{multirow}
\usepackage{multicol}
\usepackage{caption}
\usepackage{subcaption}
\usepackage{url}

\usepackage[colorlinks=true,linkcolor=blue,citecolor=blue,pagebackref=true]{hyperref}   

\begin{document}


\title{Intelligent Vision-based Autonomous Ship Landing of VTOL UAVs}



\author{
  \begin{tabular}{cc}
    \shortstack{\textbf{Bochan Lee} \\ Unmanned Weapon Systems Program Manager \\ S.Korean Navy \\
Gyeryong-si, Chungcheongnamdo, S.Korea}
 &
   \shortstack{\textbf{Vishnu Saj}\\ Graduate Student \\ Texas A\&M University \\
College Station, TX, USA}\\
    \\
   \shortstack{\textbf{Dileep Kalathil}\\ Assistant Professor \\ Texas A\&M University \\
College Station, TX, USA}
 &
   \shortstack{\textbf{Moble Benedict}\\ Associate Professor \\ Texas A\&M University \\
College Station, TX, USA}
\end{tabular}
    }
    
\maketitle

\vspace{-1cm}

\begin{abstract}

The paper discusses an intelligent vision-based control solution for autonomous tracking and landing of Vertical Take-Off and Landing (VTOL) capable Unmanned Aerial Vehicles (UAVs) on ships without utilizing GPS signal. The central idea involves automating the Navy helicopter ship landing procedure where the pilot utilizes the ship as the visual reference for long-range tracking; however, refers to a standardized visual cue installed on most Navy ships called the "horizon bar" for the final approach and landing phases. This idea is implemented using a uniquely designed nonlinear controller integrated with machine vision. The vision system utilizes machine learning based object detection for long-range ship tracking and classical computer vision for the estimation of aircraft relative position and orientation utilizing the horizon bar during the final approach and landing phases. The nonlinear controller operates based on the information estimated by the vision system and has demonstrated robust tracking performance even in the presence of uncertainties. The developed autonomous ship landing system was implemented on a quad-rotor UAV equipped with an onboard camera, and approach and landing were successfully demonstrated on a moving deck, which imitates realistic ship deck motions. Extensive simulations and flight tests were conducted to demonstrate vertical landing safety, tracking capability, and landing accuracy. \textcolor{blue}{The video of the real-world experiments and demonstrations is available at this} \href{https://www.youtube.com/watch?v=PYAO4YPIAdM}{URL}.
\end{abstract}

\begin{nomenclature}[$C_{p}$]

\nomenentry{$b$} {Close-range controller constant}
\nomenentry{$c_u$}  {Corner position-column}
\nomenentry{$c_v$}  {Corner position-row}   
\nomenentry{$d$} {Deviation from target, meter}
\nomenentry{$de(t_k)$} {Error difference between time $t$ and $t_{k-1}$}
\nomenentry{${dt}_k$} {Time difference between time $t$ and $t_{k-1}$}
\nomenentry{$e(t_k)$} {Error at time $t_k$}
\nomenentry{$K_D$} {Derivative gain}
\nomenentry{$K_I$} {Integral gain}
\nomenentry{$K_P$} {Proportional gain}
\nomenentry{$m$} {Long-range controller constant}
\nomenentry{$u$}  {Image pixel position-column}
\nomenentry{$u(t_k)$} {Control law}
\nomenentry{$v$}  {Image pixel position-row}
\nomenentry{$v_x$} {Forward relative speed, m/s}
\nomenentry{$v_y$} {Sideward relative speed, m/s}
\nomenentry{$x$} {Forward relative distance, meter}
\nomenentry{$y$} {Sideward relative distance, meter}
\nomenentry{$\nabla f(u,v)$} {Image Gradient}
\end{nomenclature}

\section{Introduction}

Landing a helicopter on a small ship at rough sea states is an extremely challenging task even for human pilots due to the small landing space, six degrees of freedom ship deck motions, limited visual references for pilots, and lack of alternative landing spots. There have been many studies in the past that focused on automating helicopter ship landing by utilizing a wide array of sensors such as GPS, vision sensors, motion sensors, LIDAR, etc. This paper investigates a novel solution that falls under the category of vision-based control system that does not use GPS signals and thus ensures its functionality in GPS-denied/spoofed environments.

Previous efforts toward autonomous ship landing involve a common process that is to estimate or track ship deck motions first, and then control the aircraft attitude to match the ship motions for landing. In order to extract the ship deck motion information, various methods have been introduced such as tracking H landing marking \cite{sanchez2014approach}, T landing marking \cite{xu2009research}, points dispersed on the deck \cite{truong2016vision}, lights \cite{holmes2016autonomous}, and infrared cooperated targets on a ship \cite{meng2019visual,yakimenko2002unmanned}. These vision-based methods have shown the limited capability for autonomous UAVs landing on ships undergoing significant motions representative of rough sea states. Fundamentally, a method that involves visual tracking of deck motion is not ideal for Vertical Take-Off and Landing (VTOL) capable Unmanned Aerial Vehicles (UAVs) that approach a ship horizontally at low altitudes because its application range is limited to the vertical space where the deck can be captured. Also, actively controlling the UAV to match the complex deck motions could excite unstable UAV attitude dynamics. This is even more unsafe when the aircraft is in close proximity to the moving deck because even a small control error can cause a catastrophic accident due to an impact by the deck. The pilot control activity and simulation of helicopter shipboard launch and recovery operations are discussed in \cite{lee2003simulation,lee2005simulation}. Furthermore, none of the previous methods were based on the Navy helicopter ship landing procedure that was established and successfully executed by pilots for decades. In fact, visually tracking the moving landing deck is exactly the opposite of what Navy helicopter pilots are trained to do. 

\begin{wrapfigure}{r}{0.42\textwidth}
\centering
\includegraphics[width=0.4\textwidth]{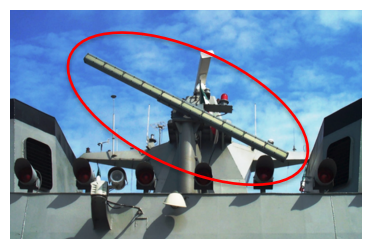}
\caption{Horizon reference bar}
\label{hrb}
\end{wrapfigure}

The present landing method is developed based on an in-depth understanding of the Navy helicopter ship landing procedure, which is discussed in Refs. \cite{tuttle1976study, lumsden1998challenges,soneson2016simulation,colwell2002maritime,minotra2020analysis}. Contrary to intuition, Navy pilots are trained not to follow ship deck motions for two main reasons. First, spatial disorientation can occur when a pilot has no fixed, visible horizon to refer to, which is a critical element for maintaining a proper sense of helicopter attitude independent of ship motions. The key visual aid that helps pilots to land safely is a "horizon reference bar" shown in Fig. \ref{hrb}, which is gyro-stabilized to indicate a perfect horizon regardless of ship motions and is widely used in most modern Navies \cite{stingl1970vtol,nato}. Thus, pilots can land a helicopter by referencing the horizon bar without responding to ship motions.

Second, constantly changing the helicopter attitude to match ship deck motions can trigger unstable helicopter dynamics introducing serious potential hazards. Hence, a pilot tries to control the helicopter in a stable manner independent of the ship's roll and pitch motions, and then lands vertically. It is also recommended to land quickly in order to prevent the ship deck from impacting one of the helicopter landing gear skids/wheels causing a rollover. This vertical landing manner is proven to be safe within the operating limits of the helicopter and currently being used in practice. The present technical approach towards automating such a landing procedure consists of machine vision to obtain the relative position and heading of the aircraft and a control system to execute the approach and landing maneuvers. 

The vision system is hybrid in nature with two different methods, a machine learning object detection and a classical computer vision method, each of which is designed to operate depending on the relative distance to the landing pad. In the long-distance, the machine learning object detection method is applied to identify the landing platform (the ship), and an image-based control is utilized in the autonomous flight control system. In recent years, there have been many studies to develop algorithms that guarantee fast detection as well as higher accuracy, which are essential to reliable UAV operations. Various algorithms and architectures of Convolutional Neural Network (CNN), a class of deep neural networks, have been proposed such as Region-based CNN (R-CNN) \cite{girshick2014rich,girshick2015fast,ren2016faster}, Single Shot Detector (SSD) \cite{liu2016ssd}, and You Look Only Once (YOLO) \cite{redmon2016you,redmon2017yolo9000,redmon2018yolov3}. R-CNN is classified as a two-stage detector that combines region-proposal algorithms with CNN to extract 2,000 regions via a selective search, then classifies the selected regions on the image. Later, its variant Faster R-CNN is introduced to improve the detection speed by replacing the slow selective region search process. Meanwhile, SSD and YOLO are classified as a one-stage detector that regards object detection as a regression problem by taking an input image and simultaneously learning the probability of an object class and bounding box coordinates. According to the studies that compared the state-of-the-art algorithms, YOLOv3 demonstrated faster detection performance than Faster R-CNN and SSD \cite{redmon2018yolov3,benjdira2019car}. Hence, the YOLOv3 algorithm is selected to train an object detector that is able to detect a ship and a horizon bar in real-time. It is not only visually tracking the object but that information is relayed in real-time to the autonomous flight control system. Once the object is detected, it provides the object position and its bounding box in the image to the autonomous flight control system. Even though the actual relative distances are not estimated, the size of the object and its position in the image are sufficient information to control a UAV to approach the ship from a long distance. The verified maximum range is approximately 250 meters (820 feet) when the object occupies an area of 1.8 x 1.8 meters (6 x 6 feet). Considering the range is proportional to the object's occupying area in the image, a typical small ship where the rear-side occupies 15 x 15 meters (50 x 50 feet) area can be detected from 17.3 kilometers (9.3 nautical miles) away. 

On the other hand, obtaining accurate relative position and orientation from a captured image is crucial for the final approach and even more important for precise landing on the ship deck. Instead of detecting an object as a whole using machine learning, computer vision techniques are applied to extract particular points of interest. To name a few, edge and line detection \cite{zishan2007computer,wang2007vision}, corner detection \cite{sharp2001vision,jihong2004method}, and contour detection \cite{lange2009vision} are previously applied to aircraft landing applications. In the present system, the visual cue to track is a horizon bar that has a rectangular shape and green color. To detect distinctive points on the bar, multiple processes such as image filtering, contour detection, corner points detection, and screening are conducted in this order. From the detected points in the 2D image, UAV relative positions and orientations are estimated using Perspective-n-Point (PnP) method. The accuracy of the present vision system to sub-centimeter and sub-degree levels have been previously demonstrated by the authors \cite{lee2020vision}.

By using the information provided by the vision system, the flight controller manipulates the UAV to approach and land. To this end, various control systems that uses vision sensors without GPS have been investigated in the literature, such as Proportional-Derivative (PD) control \cite{daly2015coordinated,holmes2016autonomous}, a Proportional-Integral-Derivative (PID) control \cite{araar2017vision,truong2016vision}, gain-scheduled PID control \cite{lee2020vision}\cite{takahashi2017autonomous}, Linear Quadratic Regulator (LQR) \cite{lee2018helicopter,lee2020development,ghamry2016real}, adaptive control \cite{hu2015fast,kim2016landing,xia2020adaptive}, discrete-time nonlinear model predictive control \cite{vlantis2015quadrotor}, and reinforcement learning based control \cite{rodriguez2019deep}. In the present system, a nonlinear control system with the Kalman filter is uniquely designed to operate accurately and robustly in the presence of time delays and sensor noise. The Kalman filter reduces the noise in estimation, however, small noise can be amplified when incorporated into the derivative controller due to the numerical differentiation process. A novel nonlinear controller is developed on top of the Kalman estimator to prevent the controller from responding to unrealistic estimations (or large fluctuations). It multiplies the estimation difference by the probability of its occurrence that follows a normal distribution. In this manner, the controller probabilistically perceives if the estimation is physically possible or not and then determines how to respond with a control input.

To demonstrate the safety of the vertical landing maneuver, which in this case is independent of the ship motions, realistic ship motions are implemented on a six Degrees Of Freedom (DOF) motion platform. The first ship motion case is from the Oliver Hazard Perry Class FFG Frigate, which is a small ship with a single landing deck. The ship motions at the sea state of 6 that refers to a wave height of 4 to 6 meters with the wave direction of 60$^{\circ}$ are scaled down for the 1.22 x 1.22 meters (4 x 4 feet) platform and are provided in \cite{sanchez2014approach}, which are also similar to measured ship motion data presented in \cite{lawther1988motion}. The second ship motion case is the FFG 7 Class ship motion limits which are 3$^{\circ}$ of pitch and 8$^{\circ}$ of roll as defined in the Naval Air Training and Operating Procedures Standardization (NATOPS) \cite{natops}. The period of pitch and roll motions are selected as 10.1 seconds and 6.5 seconds according to the study of typical small ship motions conducted by the Sandia National Lab \cite{doerry2008ship}. Vertical landings are conducted at random instances of deck motions.

The developed autonomous system is systemically verified in every aspect through extensive simulations and flight tests. A platform is constructed to reproduce ship structure and deck motions using a 6 DOF Stewart platform. A quad-rotor-UAV Parrot ANAFI \cite{parrot} live streams a video to a base-station computer, which processes the image frames to transmit control inputs through WIFI. Indoor and outdoor flight tests demonstrated the robust autonomous tracking capability and the safety of vertical landing.

\vspace{0.2cm}
The following are the key contributions of this paper. 
\vspace{-1mm}
\begin{itemize}
  \setlength\itemsep{0em}
  \item[$\bullet$] Automated Navy helicopter ship landing procedure for VTOL UAVs and verified the safety of vertical landing in the presence of realistic and challenging ship motions.
  \vspace{-1mm}
  \item[$\bullet$] Developed a state-of-the-art machine/deep learning based vision system that can identify and track objects of interest (ship platform and horizon bar) at a long distance.
  \vspace{-1mm}
  \item[$\bullet$] Developed a nonlinear control system that can be effective for real-time autonomous flight in the presence of time delay caused by the machine/deep learning based detection.
  \vspace{-1mm}
  \item[$\bullet$] Demonstrated a fast and reliable method to extract points of interest from an image by combining classical computer vision and screening algorithms.
  \vspace{-1mm}
  \item[$\bullet$] Developed a novel nonlinear controller along with a probabilistic algorithm to prevent large incorrect control inputs due to non-physical estimations to enable robust and smooth tracking.
  \vspace{-1mm}
  \item[$\bullet$] Determined a safe landing boundary considering deck motions, UAV size, and safety margin, and demonstrate the vertical landing accuracy via flight tests.
  
\end{itemize}

The outline of the paper is as follows: In Section 2, details about the development of machine vision systems are described. In section 3, the nonlinear control systems are detailed and the performances are compared to other candidate controllers. The results of long-range tracking and vertical landing tests are presented in section 4 followed by the conclusions inferred from this study.

\vspace{-4mm}
\section{Machine Vision}

\begin{wrapfigure}{r}{0.6\textwidth}
\vspace{-2mm}
\centering
\includegraphics[width=0.6\textwidth]{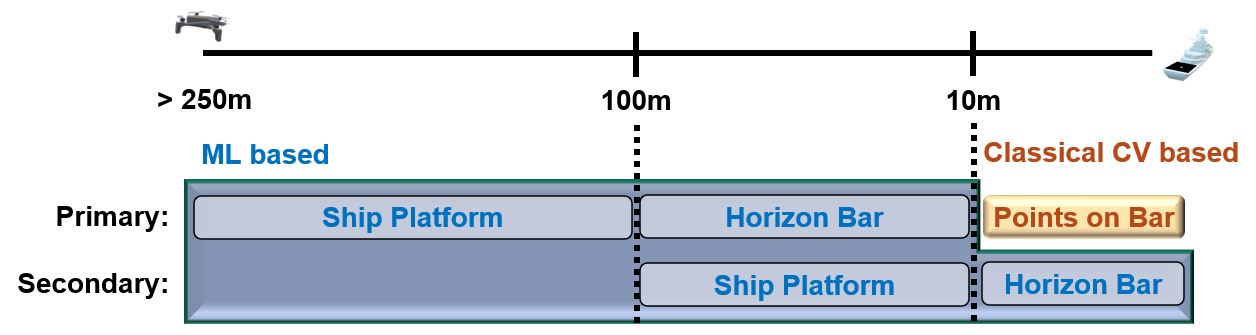}
\caption{Tracking object depending on distance}
\label{vision_over}
\end{wrapfigure}

The objective of the vision system is to provide the visual information required in a given situation. Hence, the development begins by understanding how a helicopter pilot perceives and acts during different situations while approaching and landing on a ship. First, the pilot visually confirms the ship's location from a long distance and then sets a course and speed for the approach. Second, once reaching close proximity of the ship, the horizon reference bar becomes visible, and from that point, the pilot stabilizes the helicopter by referring to the horizon bar, which remains horizontal independent of the ship motions, for a safe landing. In the present vision system, the same strategy is automated for VTOL UAVs by taking advantage of state-of-the-art machine learning object detection and classical computer vision methods. 

As shown in Fig. \ref{vision_over}, the ship, horizon bar, and corner points on the bar are used to perceive pose information depending on the distance. 

Note that it is only possible to detect the corner points at close proximity to the landing pad. The ship and horizon bar detections are used in the case when the corner points of the visual cue cannot be accurately detected.

\vspace{-0.5cm}
\subsection{Machine Learning Based Approach}
The developed machine learning object detector is engaged at long distances, where detecting the object (or ship) as a whole is of interest. Classical computer vision may achieve the same task, however, it requires explicit algorithms for detection, which leads to being complicated and also challenging to capture every aspect of the object. Instead, machine learning object detection learns the characteristics of the object thoroughly using neural networks. The speed and accuracy of this detection process are significantly improved by using Graphics Processing Units (GPUs). 

Once the object detector detects an object, it returns the position of the object and the size of its rectangle bounding box in the image. The position and size are used to determine approach course and speed, respectively. Since the detection range is proportional to the area occupied by the detected object in the image, during the early phases of approach the whole ship is detected to maximize the detection range. As the UAV gets closer to the ship, it also detects the horizon bar which provides more specific position and size information. The control system takes the information from the object detectors and takes the most appropriate control action in a given situation.

\begin{figure}[hbt!]
\centering
\begin{minipage}{0.69\textwidth}
\centering
\includegraphics[width=1\textwidth]{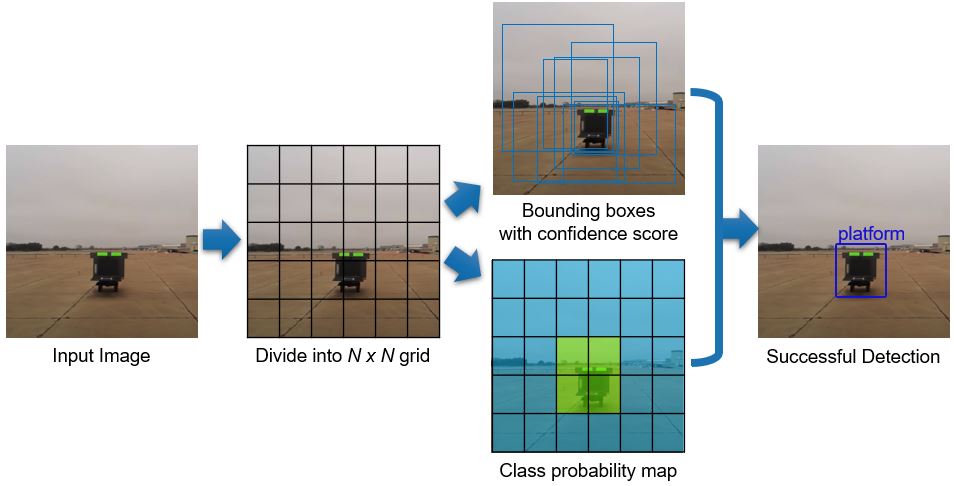}
\caption{Ship platform detection process by YOLOv3 algorithm}
\label{yolov3}       
\end{minipage}
\hfill
\begin{minipage}{0.29\textwidth}
\centering
\includegraphics[width=1\textwidth]{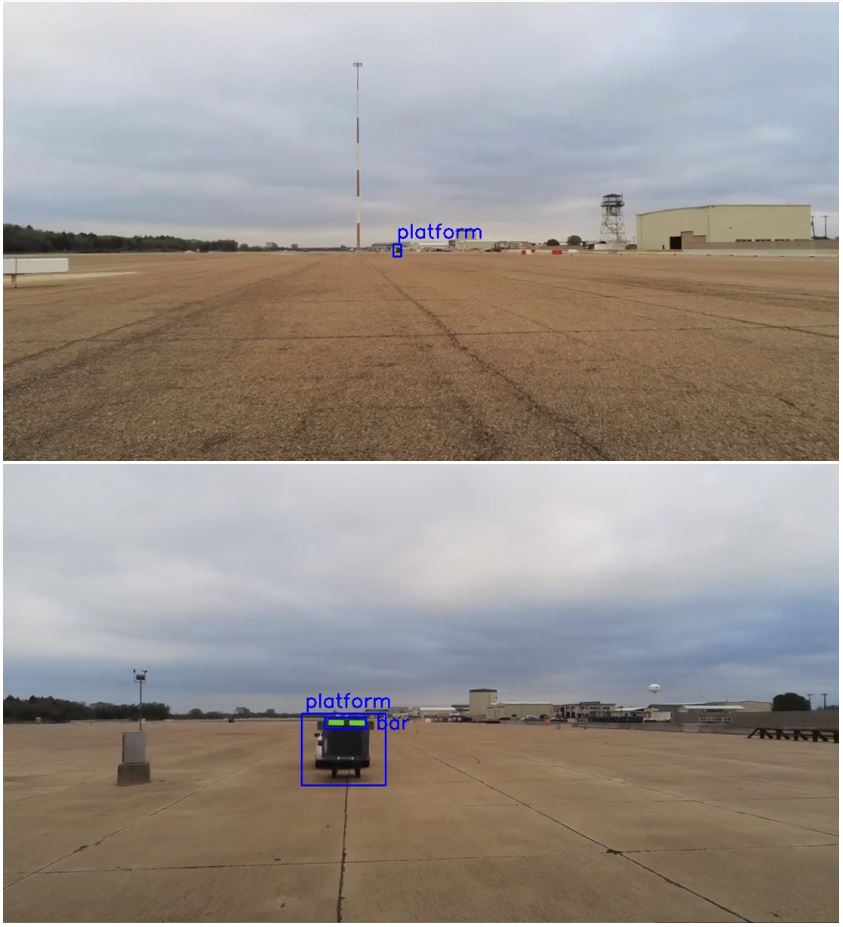}
\caption{Long and mid-range real-time object detection results}
\label{detection}
\end{minipage}
\end{figure}

Developing the object detector involves three steps, which are, collecting images, labeling objects in the images, and training the object detector. First, 2000 images of the ship platform and 1000 images of the horizon bar are collected by UAV's onboard camera. To include various object figures in training sets, the perspective, lighting, weather condition, and distance were varied as much as possible. Second, the object in the collected images is labeled by drawing a bounding box and designating the object name. This data is stored as pixel positions and the object identification number. Third, the set of labeled images are used for training via a state-of-the-art machine learning technique. In order to implement the machine learning object detection in a real-time flight system, computational speed is the first factor to consider. According to the recent studies that compare the processing time of fast detecting algorithms \cite{benjdira2019car,liu2016ssd,redmon2018yolov3}, YOLOv3 is faster than other algorithms such as SSD and Faster R-CNN while having similar accuracy in prediction. For this reason, the YOLOv3 algorithm is selected for the system. The object detection task consists of object classification and localization. Typically, an object detector is developed to detect multiple objects and classify them; however, in the present approach, two separate single object detectors are developed to detect the ship platform and horizon bar, respectively. By doing so, each detector only needs to find a particular object in the image, and the object class is automatically assigned without incurring the risk of false classification.  

The one-stage detector YOLOv3 regards the detection as a regression problem and uses a single neural network. It analyzes the entire image to predict the object-bounding box. As shown in Fig. \ref{yolov3}, the input image is divided into an N x N grid and each grid cell predicts bounding boxes with a confidence score and class probability. To better detect the object in different sizes, it predicts bounding boxes at three different scales, which helps to detect the object from a far distance. The predictions of developed detector are encoded as an $N \times N \times [3 \ast (4 + 1 + 1)]$ tensor for $N \times N$ grid cells, 3 different scales, 4 bounding box offsets, 1 object confidence score, and 1 class prediction. 

The observed maximum ranges for detecting the ship platform and horizon bar are approximately 250m and 100m, respectively. The real-time detection at different distances is shown in Fig. \ref{detection}. 

\vspace{-5mm}
\subsection{Classical Computer Vision Based Approach}

Once the UAV reaches close proximity of the ship platform, it is required to detect the visual cue and then estimate relative position and orientation for the final approach and landing. At close range, robust detection and estimation are achieved by combining computer vision techniques and developed screening algorithms. The vision system first detects the desired points of interest and then estimates the relative position and orientation. The visual cue does not need to have any particular form as long as the dimensions are known apriori. However, unique features that are distinguishable from the surroundings are favorable for detection. The current visual cue closely mimics the horizon bar on Navy ships and is installed normal to the aircraft approach course. It has two green-colored rectangles on a grey background with known size and separation as shown in the original image in Fig. {\ref{image filtering}}.


Considering the characteristics of the installed horizon bar, the corner points of the green rectangles are determined as the targets to be detected. In order to ensure robust detection and estimation in scenarios involving large UAV movements and different visibility conditions, the vision system sequentially conducts the image filtering, contour and corner detection, detected points screening, and the estimation of position and orientation.

\vspace{-0.5cm}
\subsubsection{Image Filtering}

\begin{wrapfigure}{r}{0.4\textwidth}
\vspace{-0.2cm}
\centering
\includegraphics[width=0.4\textwidth]{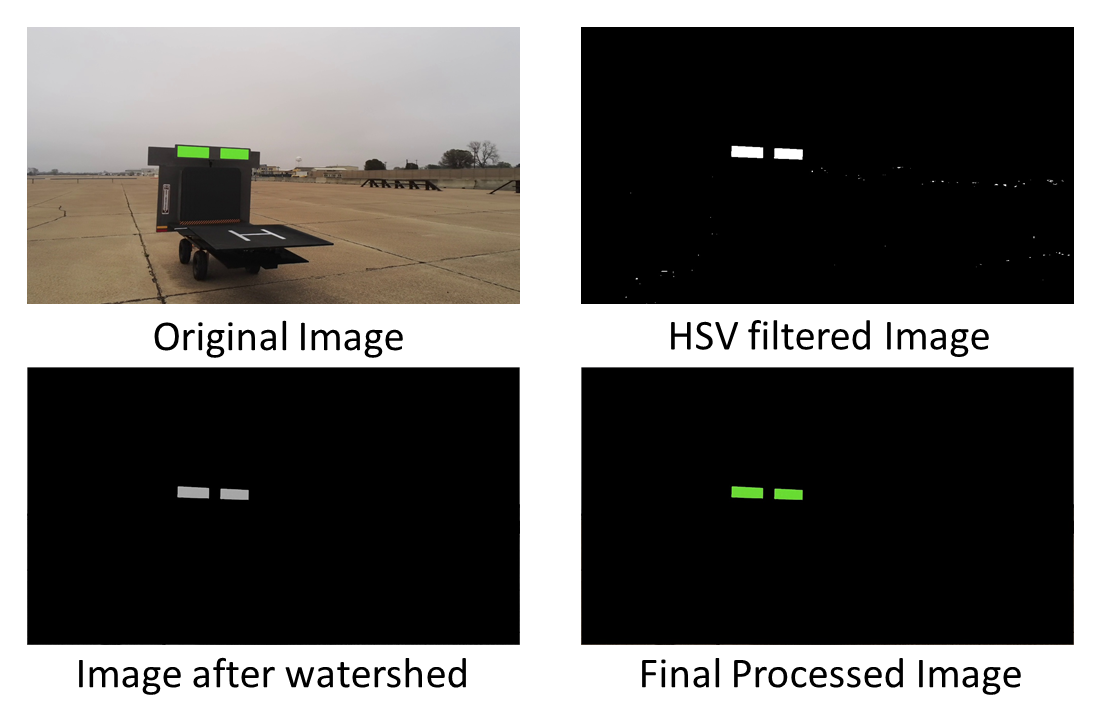}
\caption{Image filtering process}
\label{image filtering}
\end{wrapfigure}

The initial approach involves applying a Hue-Saturation-Value(HSV) filter to sort out the green rectangles (Fig. {\ref{image filtering}}). The HSV filter is preferred over Red-Green-Blue (RGB) because the RGB color space is more sensitive to light conditions. To ensure the capture of the green rectangles in different light conditions, the greater range of HSV (H: 35 - 85, S: 70 - 255, V: 90 - 255) is assigned; however, this inevitably leads to the capture of undesired portions. In the HSV filtered image, there possibly exists small white patches outside the rectangles and black voids inside the rectangles. The morphological opening technique is used to remove the white patches in the image. It first erodes an image removing any small white patches and then dilates the eroded image to preserve the original shape and size. Morphological closing is used to fill up small voids in those rectangles by dilating an image first then eroding the dilated image. By performing these operations, the rectangles are depicted as white regions on the black background. The watershed algorithm \cite{Couprie2003} is exploited to obtain clear boundaries of the rectangles (Fig. {\ref{image filtering}}). The two reference areas are obtained by eroding the processed image by 1\% and dilating the processed image by 1\%. The gap between the two areas is assumed to contain the boundaries of the rectangles. The algorithm simultaneously expands the area of the background and the rectangles toward each other until they meet at one pixel point. By connecting the points, clear boundaries of the rectangles are obtained. 

\vspace{-0.5cm}
\subsubsection{Contour and Corner Detection:}
\begin{wrapfigure}{r}{0.5\textwidth}
\centering
\vspace{-0.5cm}
\includegraphics[width=0.49\textwidth]{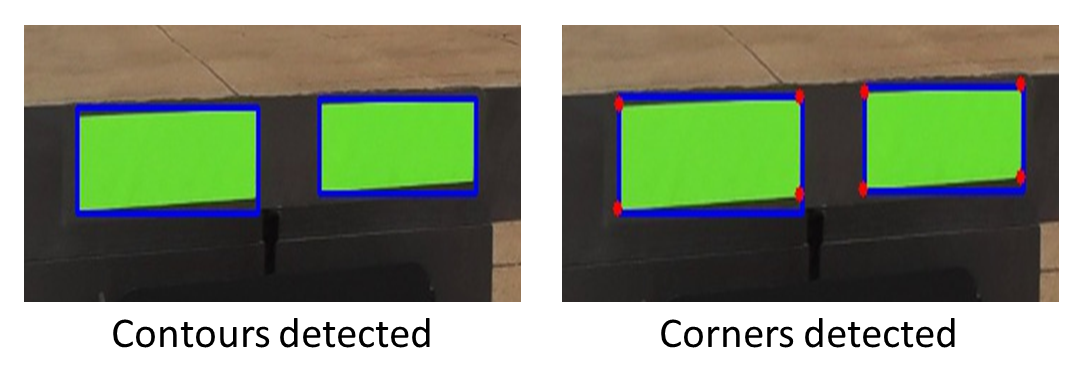}
\caption{Detection of contours and corners}
\label{corner detection}
\end{wrapfigure}
Once the green rectangles are isolated by the image filtering, the detection of contours and corner points is conducted. Even after the filtering, directly implementing any pre-existing corner detection algorithms is prone to detect some false corners. To detect the eight corners precisely, the contours of the detected region are found and bounded in rectangles first as shown in Fig. {\ref{corner detection}}. Thus, the size and shape of the detected areas are very close to the green rectangles and the corners of those bounding rectangles can be used as rough estimates of the actual corners. Second, the F\"orstner corner detection method is adopted to detect the corner points of the rectangles precisely based on the rough corners obtained by contour detection \cite{forstner1987fast}. The F\"orstner corner detection increases the accuracy by sub-pixel refinement process. A detailed decription about the F\"orstner corner detection is given in\cite{lee2020vision}

One crucial part of the F\"orstner corner detection method is the size of the window chosen for each corners in the image. It cannot be a fixed window size because the UAV is always moving towards the landing pad. The size of the detected region keeps increasing as it approaches closer to the visual cue. Hence a variable window size which is a function of the width and height is assigned as expressed in Eq. \eqref{window}:
\begin{equation} \label{window}
  \hspace{1.7cm} sw(w(t_{k}),h(t_{k})) = \frac{1}{5}\{w(t_{k}) \times h(t_{k}) \}. 
\end{equation}
Here, $sw(w(t_{k}),h(t_{k}))$ is the window size at time $t_{k}$ where $w(t_{k})$ and $h(t_{k})$ are the width and height of bounding rectangles in pixels.

\vspace{-0.5cm}
\subsubsection{Detected Points Screening}
The screening procedure is established to assure that no false corners are present. All the corners are sorted in a particular order which helps in finding the length and slope of each side of the rectangles in the image. Even though the detected regions are not perfect rectangles in the image, the width and height of the rectangles have similar lengths and slopes. A $\pm 10\%$ tolerance level is set for the lengths and a $\pm 5\%$ tolerance level is set for the slopes.


\vspace{-0.5cm}
\subsection{Relative Position and Orientation (Pose) Estimation}
The estimation is based on a single camera calibration method using a planar object \cite{est1,est2}. The geometric relation of the image and real-world coordinates can be derived using well established algorithms. The Perspective-n-Point (PnP) algorithm is implemented to determine the relative position and orientation. An iterative process is used in the PnP algorithm since it is robust for objects which consist of a planar surface. The iterative method is based on Levenberg-Marquardt optimization \cite{lev1,lev2}. In this method, the function minimizes re-projection error, which is the sum of squared distances between the observed image points and projected object points. The algorithm is detailed in \cite{lee2020vision}.   

It is apparent from multiple experiments that the yaw angle estimation becomes noisy as the camera gets farther away from the visual cue. Since the estimation is based on the number of visual cue pixels in the image, the changes in pixels with distance result in the noise. Despite this sensitivity issue, yaw estimation still shows a reasonable trend within the range that the forward relative distance is accurately estimated. To utilize the yaw estimation trend, instead of directly taking the noisy yaw angle estimation, a simple low pass filter is configured to reduce the noise level. One of the commonly used real-time data filters is Kalman filter, which predicts the current estimate by taking into account the current measured value, the previous estimate, and the noise level in the data. A single state Kalman filter was opted over other real-time filters because of the lesser number of variables required while computation. Gaussian noise is assumed to be present in the measurement data with a specified mean and variance. The underlying equations involved in estimating the yaw are:

\begin{equation}
   \begin{aligned}
     \hspace{0.0cm}CE &= PE+KG\times CM,
     \hspace{1.0cm}KG &= PrE\times\frac{1}{PrE+RN},
     \hspace{1.0cm}PrE &= PE + QN
   \end{aligned}    
\end{equation}

where CE is the current yaw estimate, PE is the previous yaw estimate, KG is the Kalman Gain, CM is the current yaw measurement, PrE is the predicted probability estimate, RN is the process noise covariance, and QN is the measurement noise covariance. RN and QN values are experimental values found by analyzing different sets of yaw angle measurements and they are set to 0.005 and 0.05, respectively. 

The estimation method for the relative position and orientation is previously validated using a Vicon motion capture system, which demonstrated sub-centimeter and sub-degree accuracy as detailed in \cite{lee2020vision}.

\vspace{-5mm}
\section{Nonlinear Control System}

\begin{wrapfigure}{r}{0.5\textwidth}
\vspace{0.5cm}
\centering
\includegraphics[width=0.49\textwidth]{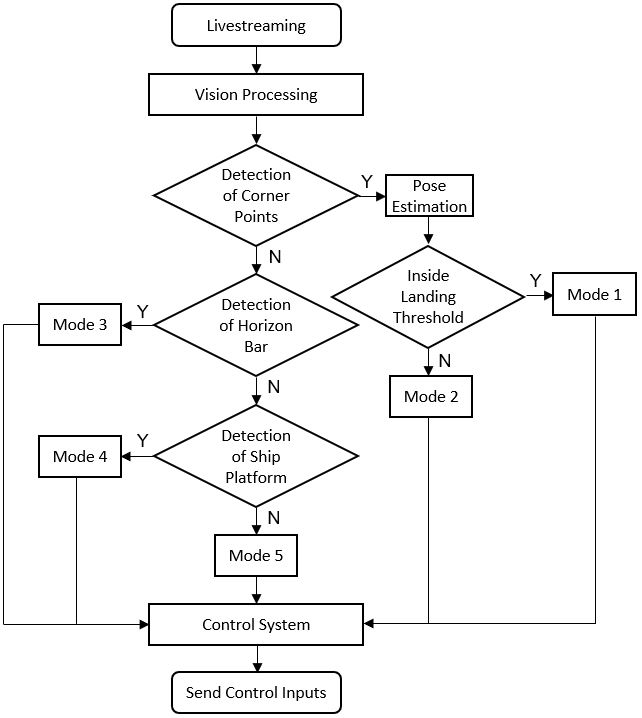}
\caption{Flowchart showing vision-based control system}
\label{flowchart}
\end{wrapfigure}

The sequential steps involved in the vision-based control system are live-streaming the video from UAV's onboard camera to an external base-station computer, image processing, feedback controller generating the control inputs, and transmitting the control inputs back to the UAV. The processing time is considerably affected by the type of detection method engaged in the vision system. The average time to complete one cycle is 0.5 seconds when the machine learning detection is engaged and 0.03 seconds when the detection of the rectangle corner points is engaged. In order to cope with the time delay as well as potential sensor and estimation noise, nonlinear controllers, which can adapt to different situations are developed. Particularly, five different flight modes are configured according to the situations perceived by the vision system as shown in Fig. \ref{flowchart}.

In the case that the computer vision system provides estimated positioning data based on corner points detection at a close distance, mode 1 triggers vertical landing command upon the satisfaction of landing condition. Mode 2 makes the UAV approach the designated landing point by engaging the probabilistic nonlinear control system. At a long distance where the corner points detection is unavailable, the machine/deep learning-based vision system provides the object area and location in the image at a long distance. In this case, the exponential nonlinear control system is primarily engaged with mode 3 if the horizon bar is detected. If not, mode 4 is engaged based on ship platform detection. The control system is designed to utilize previously fed data when no vision data is available, which is activated by mode 5.

\vspace{-5mm}
\subsection{Long-range Tracking Controller}

\begin{wraptable}{r}{0.65\textwidth}
\centering
\caption{Selected constants for nonlinear exponential controller}
\vspace{2mm}
\label{tab:non_1}       
\begin{tabular}{|c|c|c|c|c|c|}
\hline
\multirow{2}{*}{Flight Mode} & \multirow{2}{*}{Controller} & \multirow{2}{*}{m} & \multirow{2}{*}{a} & \multirow{2}{*}{c} & \multirow{2}{*}{d}\\
& & & & &\\
\hline \hline
\multirow{3}{*}{Ship Platform Tracking} &  pitch & 0.008 & 0 & 5000 & 0\\ 
                                        & roll & 1.2 & 0.0158 & 640 & 1\\ 
                                        &  heave & 3.0 & 0.0108 & 360 & 1\\

                               \hline
\multirow{3}{*}{Bar Tracking} &  pitch & 0.004 & 0 & 3400 & 0\\ 
                              &  roll & 1.0 & 0.0158 & 640 & 1\\ 
                              &  heave & 5.0 & 0.0108 & 360 & 1\\ 

\hline
\end{tabular}
\end{wraptable}
In the long-range where the entire landing platform is detected as a whole, the information provided by the vision system is the platform size and position in the camera view. The flight control system in this region deals with a relatively slow update rate that is 0.5 seconds on average. When the update rate is fast enough, the discrete system that receives sensor data at each update time can be a good approximation to the continuous system. Thus, integral and derivative controllers can be configured by using the summation and difference of error. However, the summation and difference are not good approximations for integral and derivative controllers when the update rate is slow. Hence, the nonlinear controller is designed to achieve the control task in the presence of the slow update rate by applying the nonlinear exponential gain $K_{P}\{e(t_{k})\}$ as shown in Eq. \eqref{non1}.

\begin{equation}
\label{non1}
\begin{aligned}
e(t_{k}) \hspace{0.1cm}=& \hspace{0.1cm}r(t_{k}) - c\\
u(t_{k}) \hspace{0.1cm}=& \hspace{0.1cm}K_{P}\big\{e(t_{k})\big\} \times e(t_{k}) = \left \lbrace \begin{matrix} - m(e^{ae(t_{k})} - d) \times e(t_{k}) \hspace{0.3cm} (e(t_{k}) < 0)\\ \hspace{0.25cm} m(e^{ae(t_{k})} - d) \times e(t_{k}) \hspace{0.3cm} (e(t_{k}) \geq 0)  \end{matrix}  \right.
\end{aligned}
\end{equation}

In Eq.~\eqref{non1}, $e(t_{k})$ denotes the error at time $t_{k}$, $r(t)$ is the relative position at time $t_{k}$, and $c$ is the setpoint. The control law, $u(t_{k})$, has exponential term in the nonlinear gain $K_{P}\{e(t_{k})\}$ to decay control magnitude exponentially near zero error. The constants $m$, $a$, $c$, and $d$ selected for the ship platform and bar tracking are provided in Table \ref{tab:non_1}.

\begin{figure}[hbt!]
\centering
\begin{minipage}{0.49\textwidth}
\centering
\vspace{-0.0cm}
\includegraphics[width=1.0\textwidth]{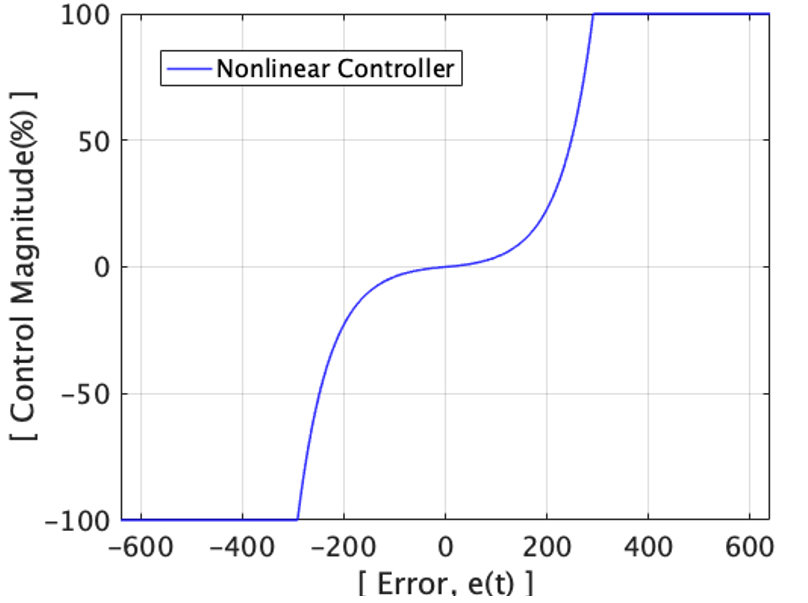}
\caption{Exponential nonlinear roll control input}
\label{expnon}
\end{minipage}
\hfill
\begin{minipage}{0.49\textwidth}
\centering
\vspace{-0.0cm}
\includegraphics[width=1.0\textwidth]{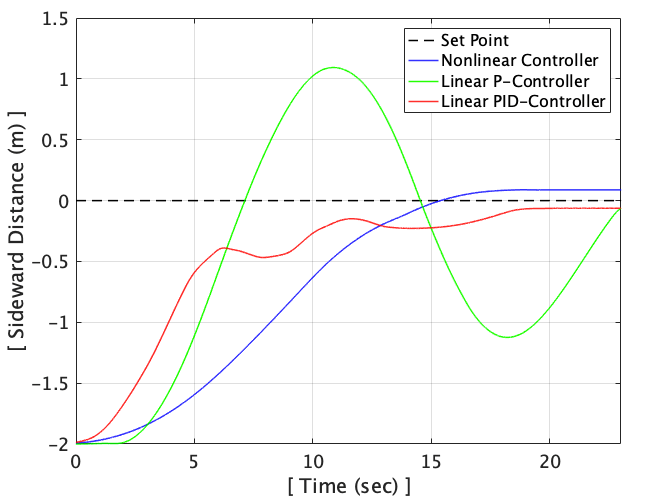}
\caption{Effect of exponential nonlinear Controller}
\label{expconresult}
\end{minipage}
\end{figure}

The desired object size, the image center position in the horizontal direction, and the image center position in the vertical direction are the setpoints for the pitch, roll, and heave controllers, respectively. Unlike the roll and heave controllers, the pitch controller with the assigned parameters returns to a linear proportional controller, which is sufficient to approach the ship in the long-range. There is also a yaw controller that can control the heading angle to set the approach course. It is designed as a nonlinear probabilistic control system, which is the same as the close-range tracking yaw controller and detailed in the following section. The only difference is that a magnetometer is used to read the current heading in the long-range and the vision system provides the estimated heading angle during close-range tracking. The nonlinear roll control input is shown in Fig. \ref{expnon}.

Depending on the constant value $a$, the rate of change of control magnitude varies. In the roll controller case, constant $a$ is selected as 0.0158. The maximum and minimum control magnitudes are limited to 100 and -100, respectively. Even though it utilizes only the error at time $t$, it can minimize the overshoot around the setpoint where the error is zero by decaying quickly. On contrary, it yields a large control magnitude as the error becomes larger.  

In the case of a slow update rate, the nonlinear controller is able to achieve stable setpoint tracking. The effect of the nonlinear roll controller is compared to the conventional linear proportional controller and PID controller as shown in Fig. \ref{expconresult}.

It demonstrates that the nonlinear controller can prevent the system from overshooting the setpoint. However, the linear proportional controller cannot stabilize the oscillations because it computes the control input by multiplying the error with a fixed gain value. Even the linear PID controller is not able to stabilize the oscillations in an effective fashion due to the slow update rate. 

However, as seen from Fig. \ref{expconresult}, the nonlinear controller has some steady-state error. Considering the long distances at which the nonlinear controller is engaged, the steady-state error is not an issue because this is not the final error of the entire approach and landing maneuver, but the initial error for the corner tracking system, which takes over the control at close distances. Therefore, it is more imperative to control the aircraft stably than to achieve the minimum steady-state error while the UAV is approaching from a long distance.

\subsection{Close-range Tracking Controller}

At close range, when the vision system can reliably detect the corners of the rectangles on the visual cue, the vision-based controller utilizes the estimated relative position and orientation data to yield control inputs. The average update rate is 0.03 seconds, which is significantly faster than the machine learning object detection that is applied at long distances. Having integral and derivative controllers along with the proportional controller enables precise tracking. Even though the estimation provides accurate position and orientation data with sub-centimeter and sub-degree error, a small error difference between subsequent time steps can yield large and noisy control magnitude since it depends on the difference in error divided by the small update rate of 0.03 seconds. To take advantage of the derivative controller with minimum noise, the Kalman filter and nonlinear derivative controller are designed. 

A single state Kalman filter is applied with the unity feedback, which means that it does not require the prediction from the model. This model-free estimator reduces the noise, however, it uses error difference values for estimation. Therefore, it will be affected by incorrect and unrealistic error difference values that may occur from time to time. To reject this intermittent unrealistic estimation effectively, a novel nonlinear derivative controller with linear proportional and integral controllers is designed by utilizing the normal distribution concept as described in Eq. \eqref{non2}. 

\begin{equation}
\label{non2}
\begin{aligned}
de(t_{k}) \hspace{0.01cm}=& \hspace{0.01cm}e(t_{k}) - e(t_{k-1})\\
u(t_{k}) \hspace{0.01cm}=& \hspace{0.01cm}K_{P}e(t_{k}) + K_{I}\sum_{k=1}^{k}e(t_{k})dt_{k} + K_{D}\big\{de(t_{k})\big\}\frac{de(t_{k})}{dt_{k}} 
\end{aligned}
\end{equation}

$de(t_{k})$ is an error difference between time $t_{k}$ and $t_{k-1}$, and $u(t_{k})$ is a control law that has linear proportional and integral terms as well as the nonlinear derivative term. $K_{P}$ and $K_{I}$ are constant proportional and integral gains, and $K_{D}\{de(t_{k})\}$ is the nonlinear derivative gain that is a function of error difference, $de(t_{k})$, as specified in Eq.~\eqref{non3}.

\begin{equation}
\label{non3}
\begin{aligned}
\hspace{2.2cm} f(x) \hspace{0.1cm} =& \frac{1}{\sqrt{2\pi\sigma^{2}}} e^{-0.5\frac{(x-\mu)^{2}}{\sigma^{2}}}, \quad
K_{D}\big\{de(t_{k})\big\} \hspace{0.1cm}=& \hspace{0.1cm} be^{-0.5\frac{(de(t_{k})-\mu)^{2}}{\sigma^{2}}}
\end{aligned}
\end{equation}

\begin{figure}[hbt!]
\centering
\begin{minipage}{0.49\textwidth}
\centering
\vspace{0.0cm}
\includegraphics[width=0.9\textwidth]{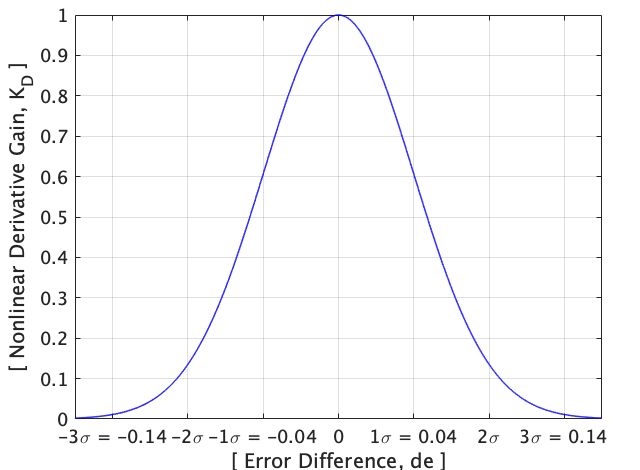}
\caption{Probabilistic nonlinear derivative controller gain}
\label{noncon2}
\end{minipage}
\hfill
\begin{minipage}{0.49\textwidth}
\centering
\includegraphics[width=0.9\textwidth]{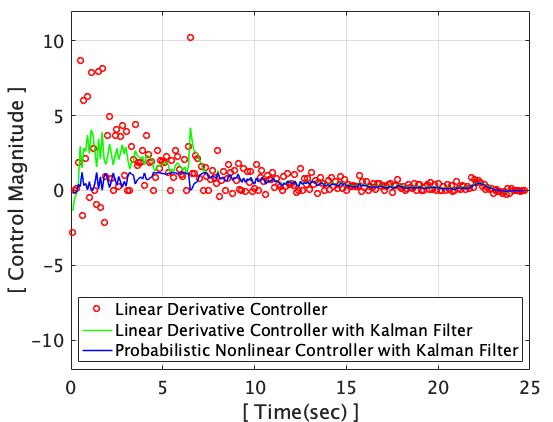}
\caption{Effect of Kalman filter and probabilistic nonlinear derivative controller}
\label{noncon2result}
\end{minipage}
\end{figure}

$f(x)$ is the probability density function that forms a normal distribution. $\sigma$ denotes the standard deviation and $\mu$ denotes the mean value. The area under the function indicates the probability that a certain range of deviation occurs. The probabilistic nonlinear derivative controller is constructed by taking the exponential term from the probability density function and multiplying constant $b$ that determines the control magnitude. Based on the observation of aircraft movement, $\sigma$ is determined as 0.04, which means the distance that the aircraft can move during 0.03 seconds has a 68.2 percent chance of being within 4 cm and a 95.4 percent chance of being within 8 cm. When $b$ is 1 and $\mu$ is 0, the probabilistic nonlinear derivative controller computes derivative gains as shown in Fig. \ref{noncon2}.\\

Depending on the error difference, $de(t_{k})$, the corresponding nonlinear derivative gain, $K_{D}$, is selected and multiplied by the derivative term $de(t_{k})/dt_{k}$. If the estimated error difference is too high (or unrealistic), then it takes a small $K_{D}$ value to significantly minimize the control input. In this way, the controller does not respond to the large noise in the error data, which can trigger undesired and unstable maneuvers. Also, the magnitude of gain can vary according to the selection of constant $b$ in Eq. \eqref{non3}. The effects of the Kalman filter and the probabilistic nonlinear controller are shown in Fig. \ref{noncon2result}.

\begin{wraptable}{r}{0.7\textwidth}
\centering
\caption{Selected gains for nonlinear controllers}
\vspace{2mm}
\label{tab:non_2}       
\begin{tabular}{|c|c|c|c|ccc|}
\hline
Mode &  Controller & $K_{P}$ & $K_{I}$ & \multicolumn{3}{c|}{$K_{D}$}\\ 
                      &                              &                        &                        & b & $\mu$ & $\sigma$\\ 
\hline\hline
\multirow{3}{*}{Corner Points Tracking} &  pitch & 7.5 & 0.05 & 4.5 & 0.02 & 0.04\\ 
                                        &  roll & 7.5 & 0.01 & 8.5 & 0 & 0.04\\ 
                                        &  heave & 15 & 0.01 & 7 & 0 & 0.02\\ 
                                        &  yaw & 5.5 & 0.05 & 1.75 & 0 & 5\\

\hline
\end{tabular}
\end{wraptable}
Red circles denote the control magnitude of the linear derivative controller without the Kalman filter. The high noise that occurs in the range of 0 to 10 seconds are due to the derivative term, $de(t_{k})/dt_{k}$. This term is sensitive because the error is divided by a small $dt_{k}$, which has an average value of 0.03 seconds. Thus, even a small error in estimation can be magnified in the derivative controller. The green line is the result after applying the Kalman filter to the linear derivative controller. The noise is reduced, however, it still yields large control inputs in response to the incorrect estimation values. The blue line shows the control inputs generated by the probabilistic nonlinear controller with the Kalman filter and these inputs are relatively small and insensitive to the large unrealistic error differences. At the 6.5 second mark, the error difference $de$ has a large value caused by incorrect estimation. In this case, the linear derivative controller with the Kalman filter reduced the magnitude to some extent but it is still affected by that particular spurious error value. However, the probabilistic nonlinear controller with the Kalman filter is able to screen out the wrong value, thereby minimizing the effect of incorrect error estimation. The finally gains selected through extensive flight tests and simulations are specified in Table \ref{tab:non_2}.

\section{Flight Testing}

The objectives of flight testing are to verify the long-range tracking capability, landing accuracy, and safe vertical landing on a platform with 6 DOF motions and forward translation. Extensive flight tests are conducted in realistic and challenging scenarios. First, the ship landing environment is mimicked by the landing pad via implementing the Oliver Hazard Perry Class FFG Frigate motions with helicopter ship landing limits. While the landing pad is in motion, the ship platform is also translating forward just like a real ship during a helicopter landing. During the approach and landing, the UAV makes transitions between the flight modes that engage with a particular vision and control method based on the perceived situation, and lands vertically without matching the UAV attitude to that of the pad. During outdoor flight testing the maximum wind speed experienced was around 9 m/s (17.5 knots) and the wind direction kept fluctuating over the course of these experiments. The GPS is only used to log positions during flight and not for control.   

\subsection{Experimental Setup}

\begin{figure}[hbt!]
\centering
\begin{minipage}{0.49\textwidth}
\centering
\includegraphics[width=0.9\textwidth]{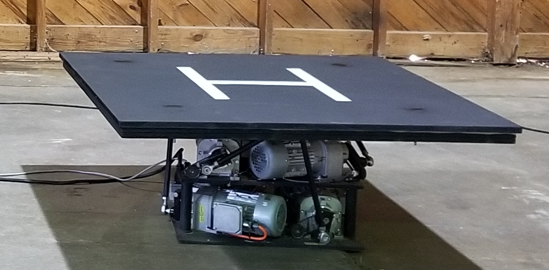}
\caption{Servos and Simulation Inc. 6 DOF motion system with 1.22 x 1.22 meters} (4 × 4 feet) Landing Deck
\label{6dof}
\end{minipage}
\hfill
\begin{minipage}{0.49\textwidth}
\centering
\vspace{-0.2cm}
\includegraphics[width=0.8\textwidth]{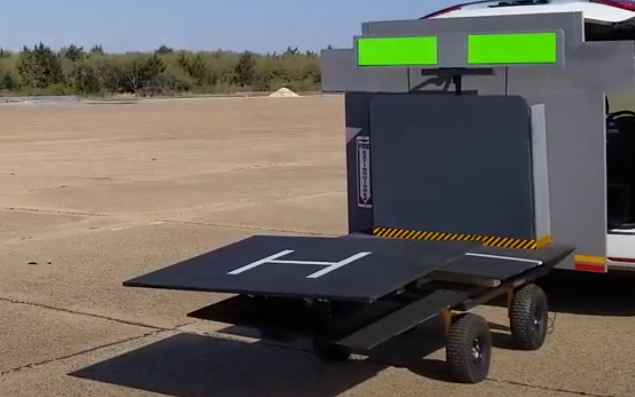}
\caption{Constructed ship platform with horizon bar and motion deck}
\label{ship}
\end{minipage}
\end{figure}

To simulate the 6 DOF motions of the ship deck, a Servos and Simulation Inc Generic Motion System (model 710-6-500-220) with a 1.22 x 1.22 meters (4 × 4 feet) landing deck as shown in Fig. \ref{6dof} is used. The ranges of roll, pitch, and yaw are ± 13$^{\circ}$, ± 15$^{\circ}$, and ± 16$^{\circ}$, respectively. The ranges of surge, sway, and heave are ± 1.02 meters, ± 1.02 meters, and ± 0.64 meters, respectively.

The ship platform is constructed including the horizon bar and motion deck as shown in Fig. \ref{ship}. The width, height, and length of the ship platform are 1.8 meters (6 feet), 1.8 meters (6 feet), and 3 meters (10 feet), respectively. The horizon bar always indicates a perfect horizon, and the motion deck has its own 6 DOF motions in addition to the forward translational motion, which is similar to what would be experienced on a real ship.

\begin{wrapfigure}{r}{0.5\textwidth}
\centering
\vspace{-0.0cm}
\includegraphics[width=0.5\textwidth]{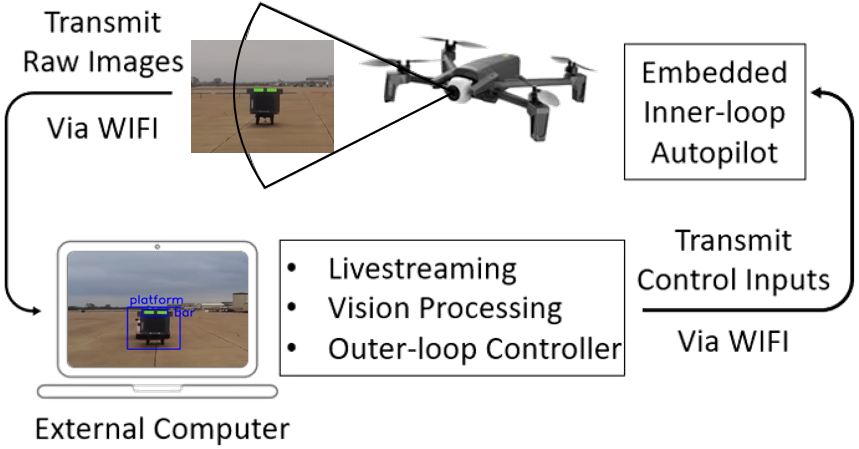}
\caption{Process of autonomous flight system}
\label{process}
\end{wrapfigure}

A Parrot ANAFI is selected as a representative VTOL UAV. It has a gimbal camera that can mechanically compensate for the roll and pitch motion of the UAV and live streams a video in 720p resolution. As shown in Fig. \ref{process}, it has the embedded inner-loop autopilot that controls each propeller's rpm to achieve the commanded inputs generated by the outer-loop vision-based control system developed in this study. 

The system used for processing is LENOVO Legion Y740-15IRH, which is composed of Intel(R) Core(TM) i7-9750H CPU @ 2.60GHz, 6 Cores, and 12 Logical Processors. It features an integrated NVIDIA GeForce GTX 1660 Ti 6GB Graphics and 8GB of LPDDR4 memory with a 128-bit interface. The Ubuntu 18.04 with Nvidia driver version 440 and CUDA version 10.2 are used. The WIFI communication is established by an external TP-LINK TL-WN722N Wireless N150 High Gain USB Adapter.

\subsection{Ship Motions and Vertical Landing Safety}

\begin{figure}[hbt!]
\centering
\begin{subfigure}[t]{0.49\textwidth}
\centering
\vspace{-0.4cm}
\includegraphics[width=1.0\textwidth]{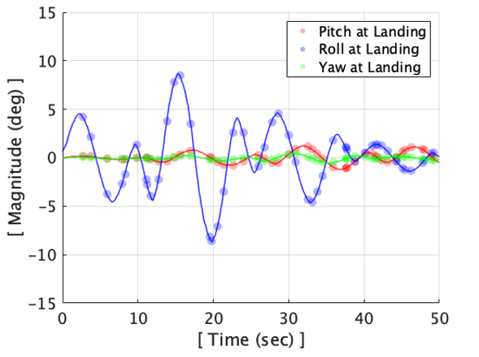}
\end{subfigure}
\hfill
\begin{subfigure}[t]{0.49\textwidth}
\centering
\vspace{-0.4cm}
\includegraphics[width=1.0\textwidth]{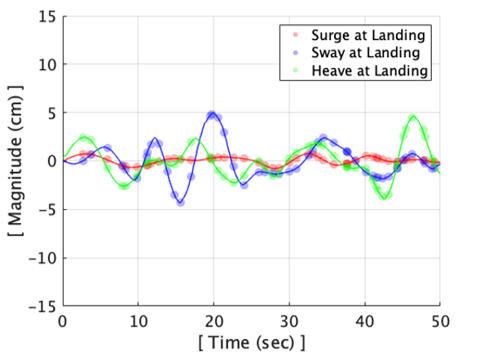}
\end{subfigure}
\caption{Oliver Hazard Perry class frigate motion and vertical landing moments}
\label{vertical_landing_1}
\end{figure}

In order to demonstrate that it is safe to land vertically without matching the UAV attitude dynamics to platform motion, landing tests are conducted while the 6 DOF platform is simulating two challenging ship motions. The first prescribed motion is the Oliver Hazard Perry Class frigate at the sea state of 6 and a wave direction of 60° introduced by Sanchez-Lopez, Jose Luis, et al. \cite{sanchez2014approach}. The frigate is 136 meters long and 14 meters wide and has a single flight deck. Sea state 6 is defined by the World Meteorological Organization (WMO) as a very rough condition that has a wave height of 4 to 6 meters \cite{davis2009beaches}. While the platform is undergoing this complicated motion, vertical landings are executed at random time instances as shown in Fig. \ref{vertical_landing_1}.

Solid red, blue, and green lines denote continuous angular (pitch, roll, and yaw) and linear (surge, sway and heave) motions of the 6 DOF platform. 50 landing tests are conducted in total and the motion of the platform at the time of each successful landing are depicted as dots. The randomly distributed landing timings demonstrate the vertical landing is safe at any moment of the Oliver Hazard Perry Class FFG frigate ship motions under the given conditions.

\begin{wrapfigure}{r}{0.4\textwidth}
\centering
\vspace{-0.4cm}
\includegraphics[width=0.4\textwidth]{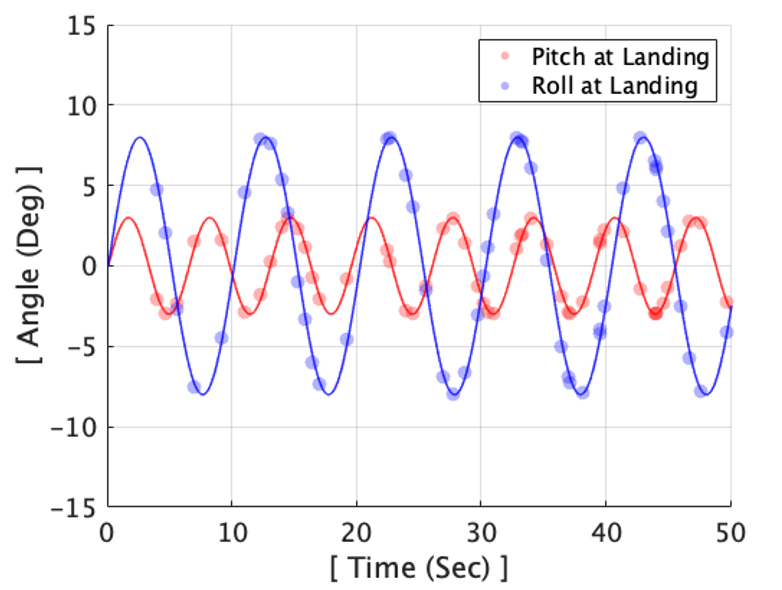}
\caption{Motions of helicopter ship landing limits and vertical landing moments}
\label{vertical_landing_2}
\end{wrapfigure}

The second prescribed motion is the Navy helicopter ship landing limits defined by NATOPS. In the case of the FFG 7 Class Ships which the Oliver Hazard Perry frigate belongs to, the ship motion limits for landing are set as ± 8° of roll and ± 3° of pitch. Even though the limits are defined by the maximum roll and pitch magnitudes, the frequency of the motion is also a crucial factor for the motions. According to the report for a similar size ship (length: 152.4 m, width: 15.2 m) motions conducted by the Sandia National Laboratory \cite{doerry2008ship}, the roll period is 10.1 seconds and the pitch period is 6.5 seconds as shown in Table \ref{tab:sandia}. 

\begin{wraptable}{r}{0.7\textwidth}
\caption{Typical ship characteristics by Sandia National Lab report}
\label{tab:sandia}       
\centering
\begin{tabular}{|ccccc|}

\hline
Ship Type & Length & Width & Roll Period & Pitch Period\\
  & (m) & (m) & (secs) & (secs)\\
\hline
Destroyer & 152.4 &15.2 & 10.1 & 6.5\\
Aircraft Carrier & 304.8 & 38.1 & 15.8 & 8.8\\
\hline
\end{tabular}
\end{wraptable}

The maximum pitch and roll magnitude with the reported periods are applied to the platform and 50 vertical landings are conducted at random moments as shown in Fig. \ref{vertical_landing_2}.

Solid red and blue lines denote continuous pitch and roll motions of the 6 DOF platform. The dots are the motions at the time of successful landings. The randomly distributed landing timings demonstrate the vertical landings are safe independent of the motion as long as it is within the operational limits.

\subsection{Tracking Capability and Landing Accuracy}

The tracking capability and landing accuracy of the developed vision-based autonomous flight control system are verified in challenging situations such as random initial positions, maximum distance of 250 meters, realistic ship motions, communication latency, sensor noise, low visibility, and windy conditions. In this fully autonomous vision-based system, GPS and a magnetometer are used only to log positions and heading angles and not for control. During flight testing, the landing pad is mimicking the ship deck motions that are introduced in the previous section.

\begin{figure}[hbt!]
\centering
\begin{subfigure}{0.49\textwidth}
\centering
\includegraphics[width=1.0\textwidth]{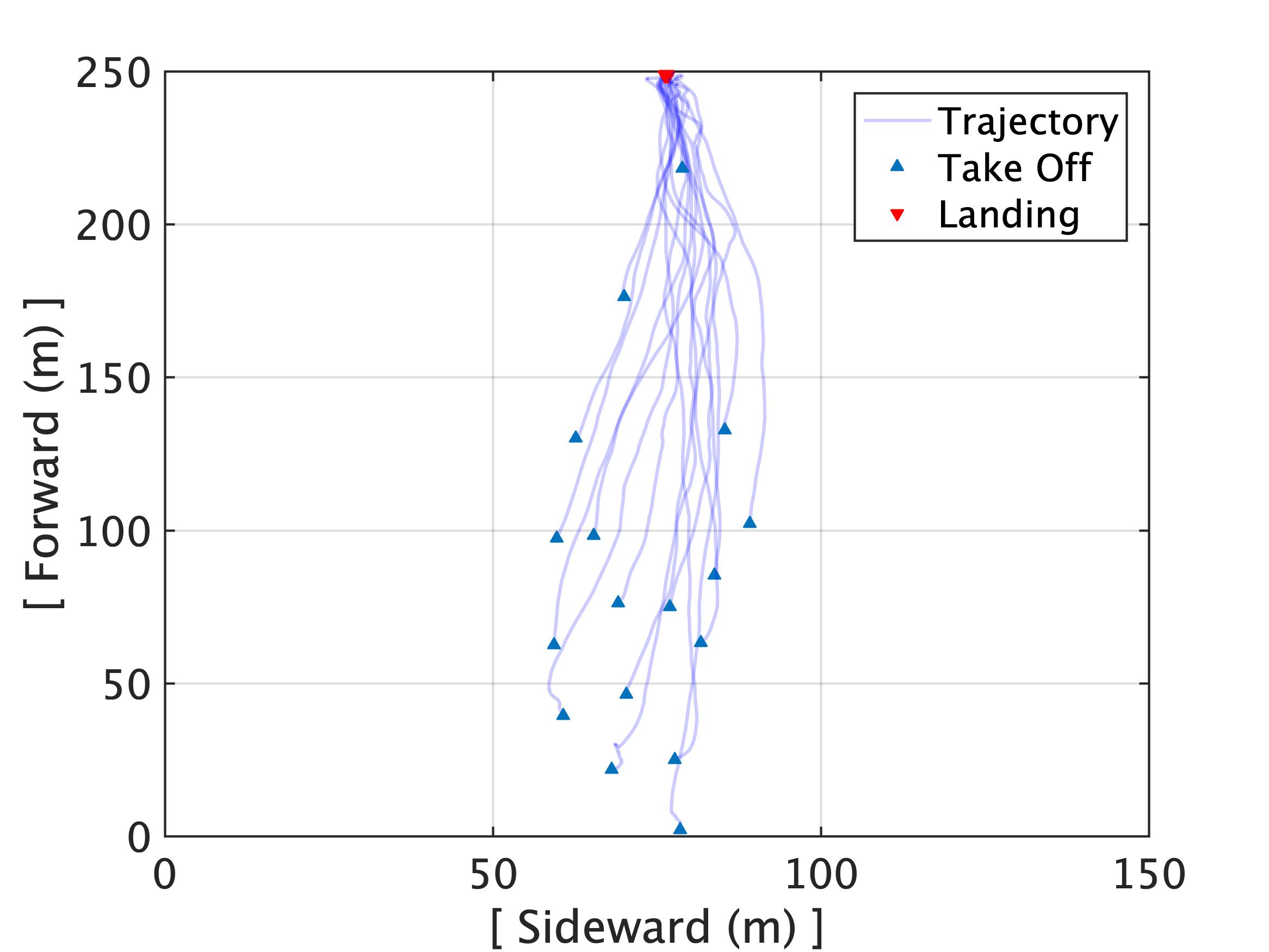}
\caption{Trajectories of long-range tracking}
\label{traj1}
\vspace{0mm}
\end{subfigure}
\hfill
\begin{subfigure}{0.49\textwidth}
\centering
\includegraphics[width=1.0\textwidth]{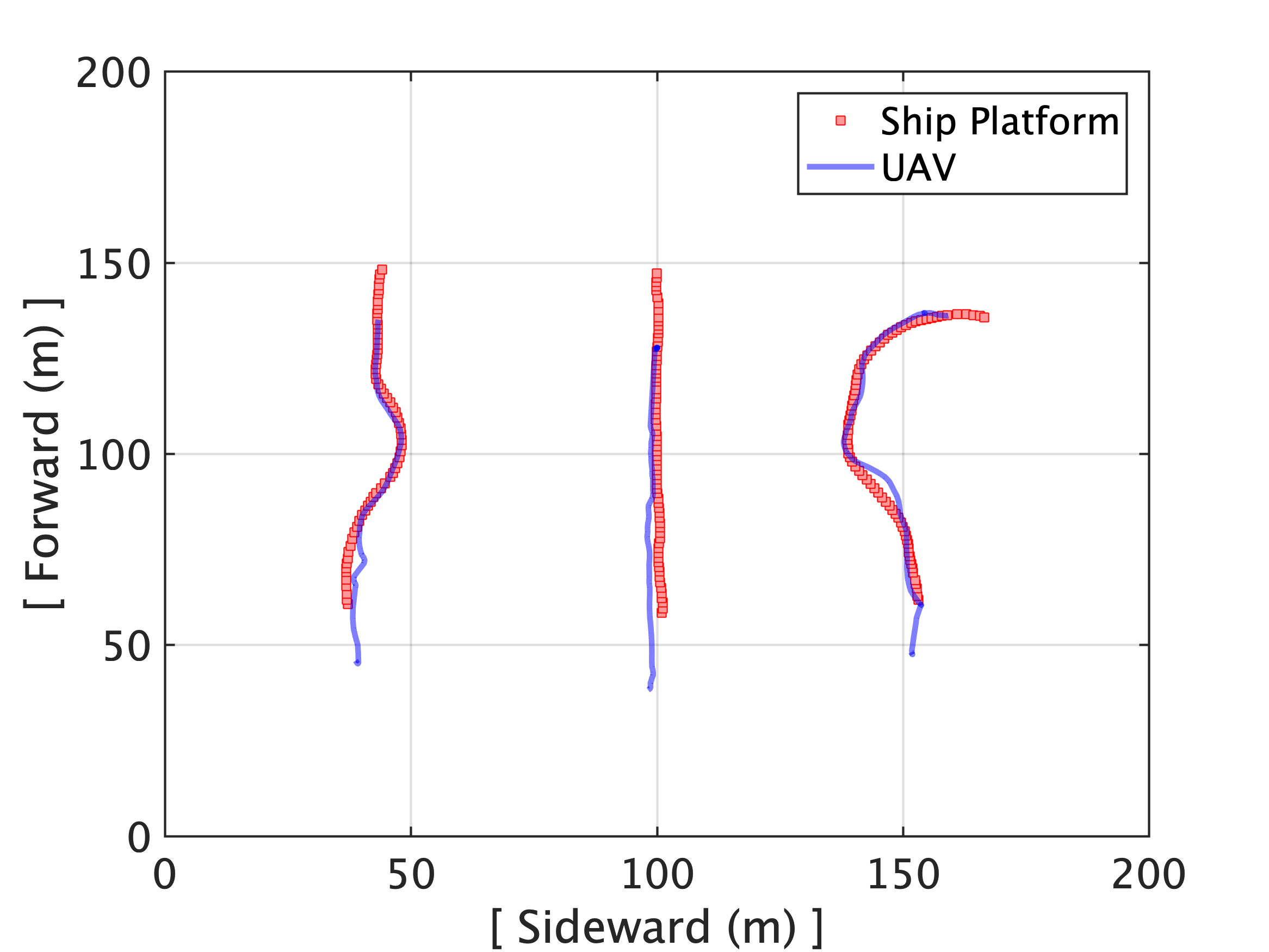}
\caption{Trajectories of moving ship tracking}
\label{traj2}
\end{subfigure}
\caption{Trajectories in different cases}
\vspace{-0.0cm}
\end{figure}

First, flight tests are conducted to verify the maximum range and smooth transition between flight modes. The trajectories are shown in Fig. \ref{traj1}. 

\begin{figure}[hbt!]
\centering
\vspace{-1mm}
\begin{subfigure}{0.49\textwidth}
\centering
\includegraphics[width=0.80\textwidth]{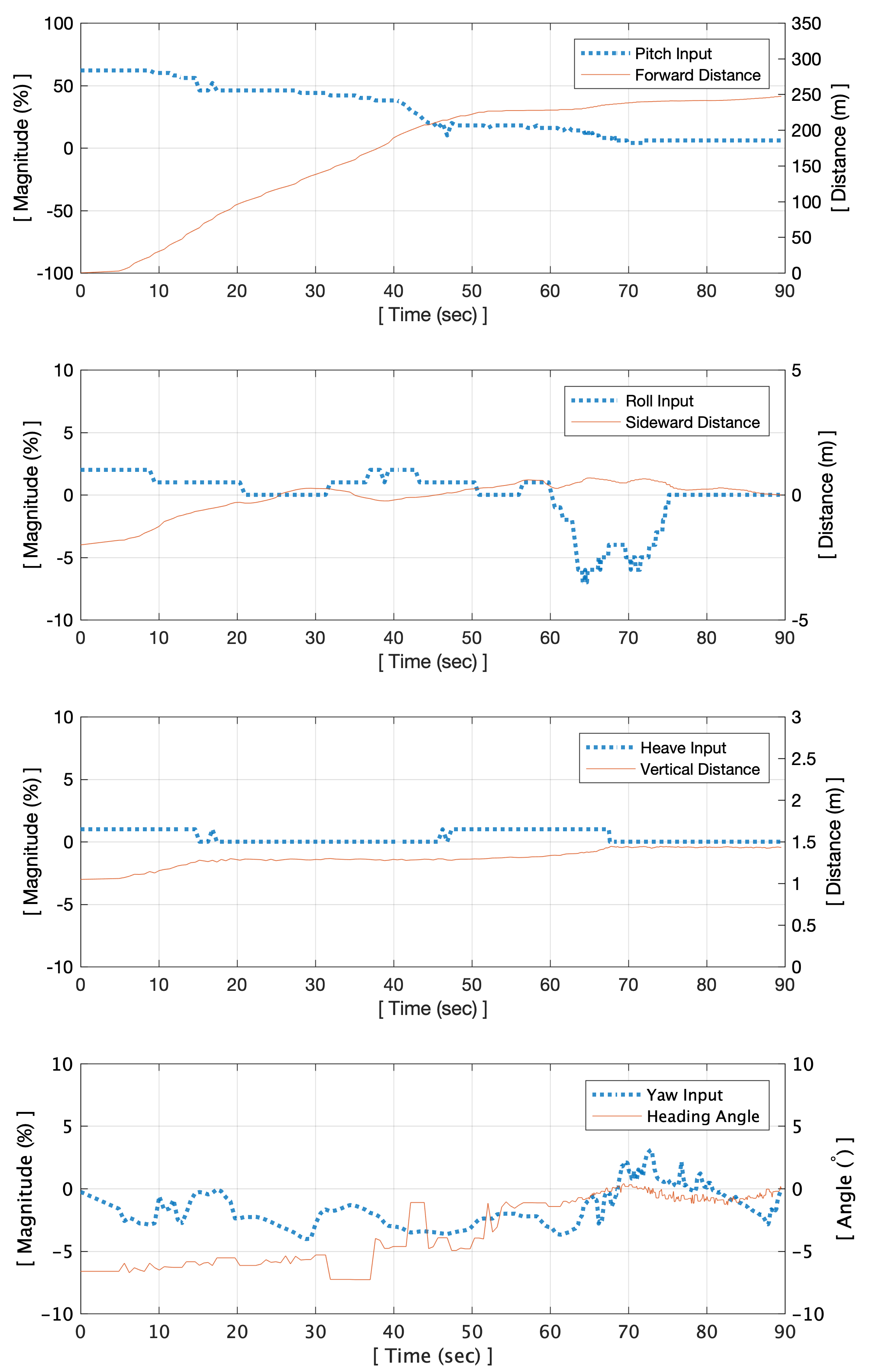}
\caption{Control inputs, distances, and heading angle during long-range tracking in time}
\label{control1}
\end{subfigure}
\hfill
\begin{subfigure}{0.49\textwidth}
\centering
\includegraphics[width=0.80\textwidth]{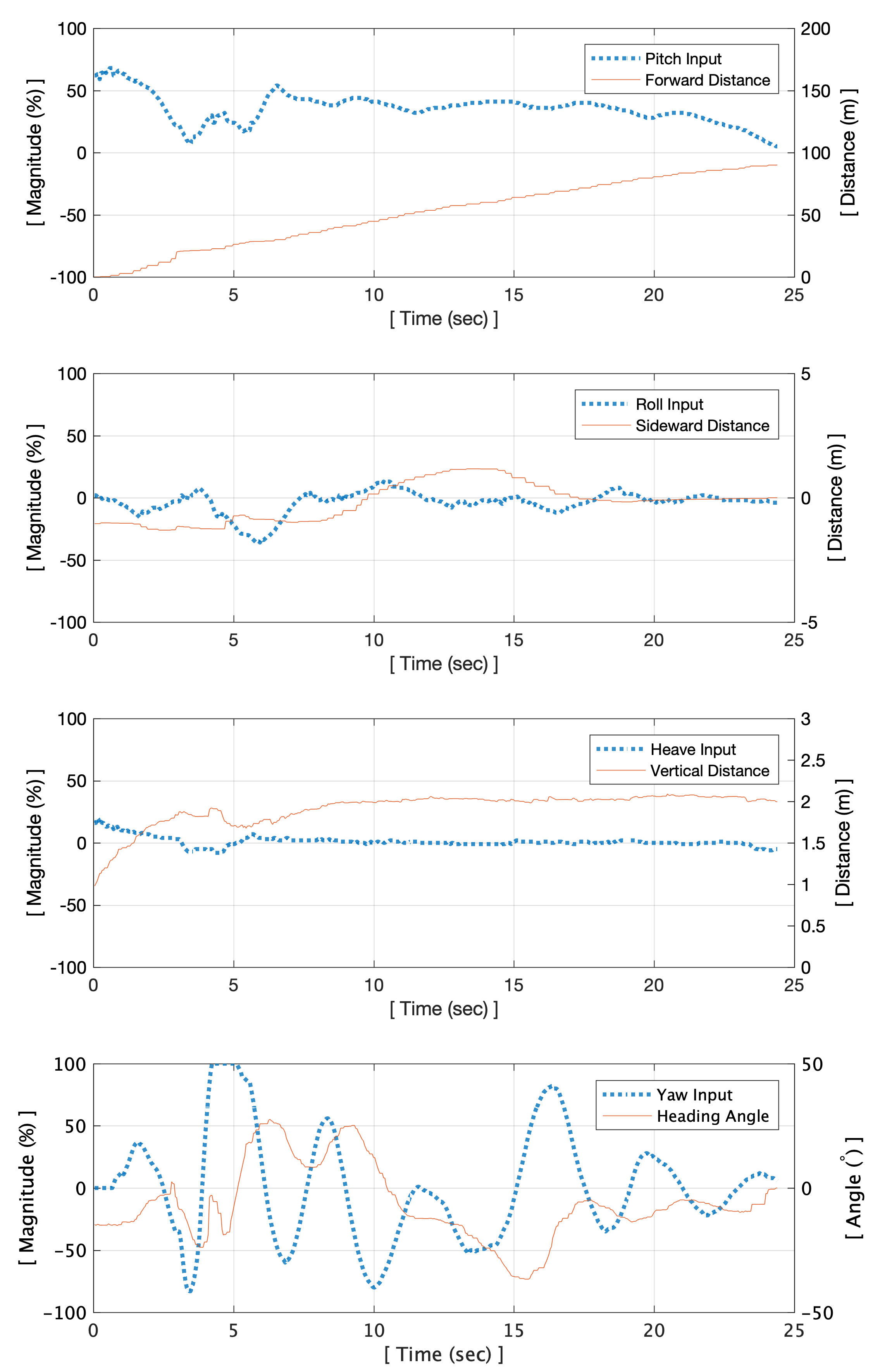}
\caption{Control inputs, distances, and heading angle of moving ship tracking in time}
\label{control2}
\end{subfigure}
\caption{Control inputs in different cases}
\end{figure}

The initial positions are widely distributed and the maximum distance between the UAV and ship platform is approximately 250 meters. During flights, the flight modes are switched from the ship platform/bar tracking to the corners tracking depending on the distance. The results demonstrate stable long-range tracking capability in the presence of wind up to 9 m/s (17 knots). The time history of control inputs for a representative case is shown in Fig. \ref{control1}. 

It shows the control inputs, distances, and heading angle after take-off until the execution of vertical landing. Each line denotes pitch, roll, heave, and yaw control inputs described as percentages that range from -100 to 100 and generated based on forward, sideward, vertical relative distance, and relative heading angle, respectively. Zero control input means neutral control input that maintains current UAV pose such as pitch, roll, altitude, and heading. While the UAV approaches the ship platform from 250 meters away, the vision-based flight control system effectively flies the UAV by smoothly switching between the flight modes from the machine learning object tracking for ship platform and horizon bar to the rectangle corner points tracking, depending on the relative distance. 

\begin{figure}[hbt!]
\centering
\vspace{-1mm}
\begin{subfigure}{0.49\textwidth}
\centering
\includegraphics[width=1.0\textwidth]{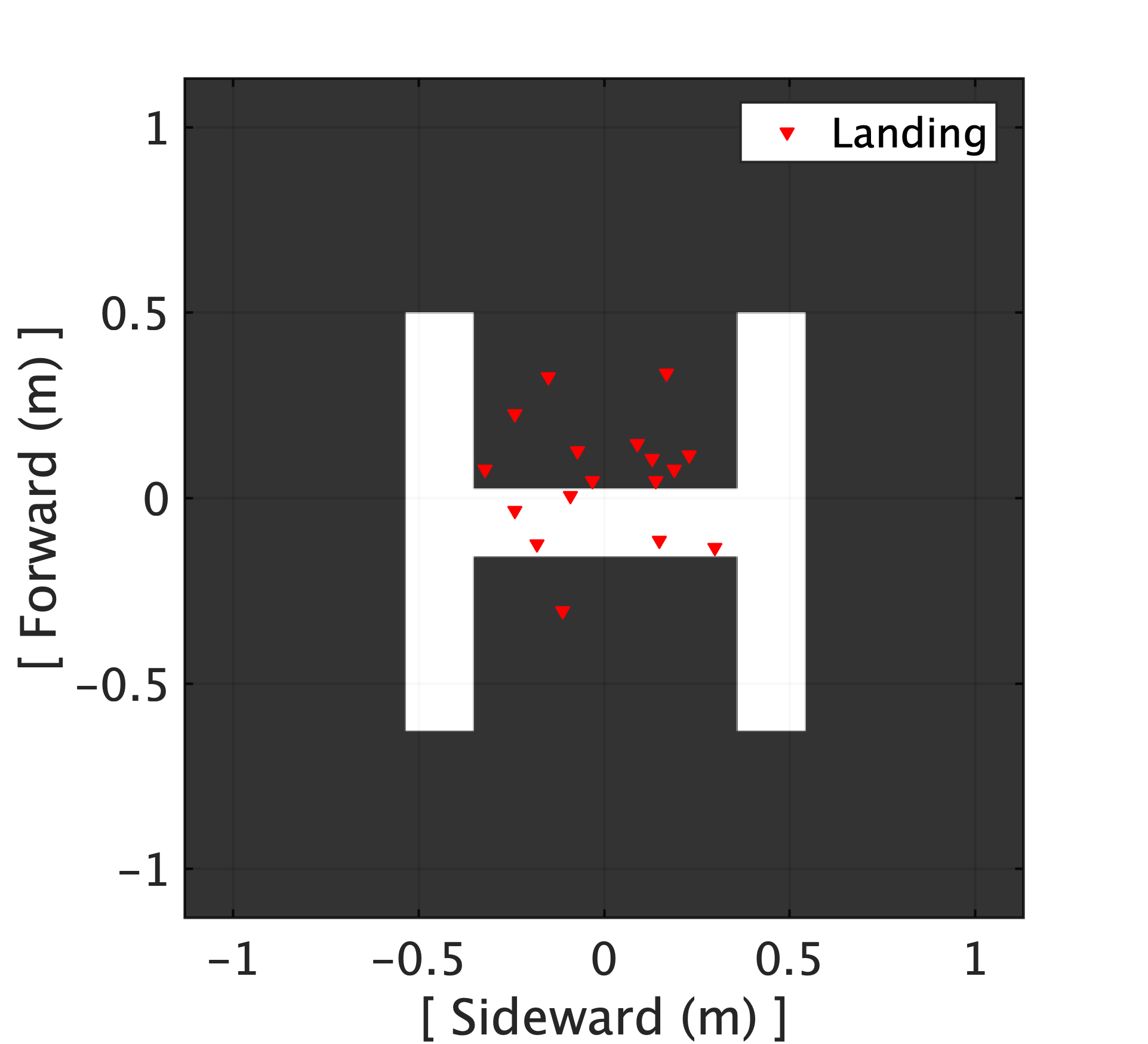}
\caption{Landing points on pad of long-Range tracking}
\label{landing_1}
\end{subfigure}
\hfill
\begin{subfigure}{0.49\textwidth}
\centering
\includegraphics[width=1.0\textwidth]{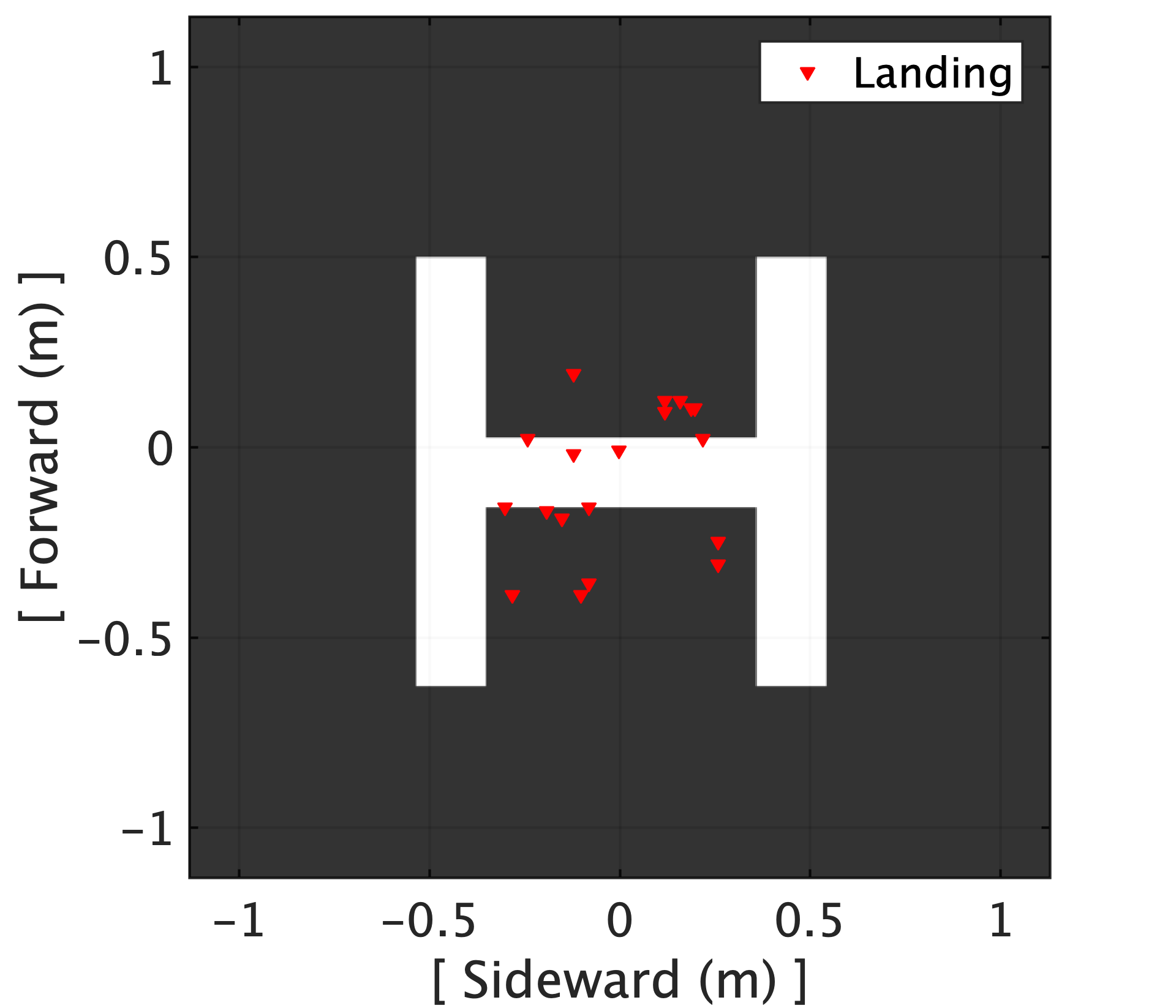}
\caption{Landing points on pad of moving ship platform}
\label{landing_2}
\end{subfigure}
\caption{Landing points in different cases}
\end{figure}

Considering the UAV and landing pad size, the safe landing threshold is set by a 0.35 x 0.35 meters square from the pad center, and once the UAV reaches the landing threshold the controller commands vertical landing. The final landing points from multiple flight tests are distributed within the set landing threshold as shown in Fig. \ref{landing_1}. 

Second, flight tests are conducted to verify the robust tracking while the ship platform is moving in different courses with its own 6 DOF motions. The three representative trajectories are shown in Fig. \ref{traj2}.

Red squares denote the trajectory of the ship platform and blue lines denote the trajectory of the UAV. It shows robust tracking while the ship platform is moving in straight, S-pattern, and 90$^{\circ}$ turn. The control inputs for S-pattern moving platform tracking are shown in Fig. \ref{control2}.

It shows the control inputs, distances, and heading angle after take-off until execution of vertical landing while the UAV is tracking the ship platform moving in an S-pattern. The pitch control input is generated to approach the ship platform that varies its speed from 0 to 4.5 m/s (10 mph). Since the ship platform changes the course abruptly up to 130$^{\circ}$, the relative sideward distance and heading also change to a great extent. Accordingly, the corresponding roll and yaw control inputs change aggressively to achieve zero relative sideward distance and heading. Vertical landings are conducted once the UAV flies into the 0.35 x 0.35 meters landing threshold while the ship platform is in motion. The final landing points from the multiple flight tests are distributed within the set landing threshold as shown in Fig. \ref{landing_2}. 


\textcolor{blue}{The video of the real-world experiments and demonstrations is available at this} \href{https://www.youtube.com/watch?v=PYAO4YPIAdM}{URL}.

\section{Conclusion}
The goal of this study is to develop an autonomous ship landing solution for VTOL UAVs by closely following the Navy helicopter ship approach and landing procedure. This automation has been achieved by using a single onboard camera without using GPS. The ship landing system consists of a machine vision system and a nonlinear control system. The machine vision system is developed while taking advantage of state-of-the-art machine learning and classical computer vision techniques. The control system is developed to generate situation-adaptive control inputs by introducing the idea of nonlinear gain variation and a probabilistic approach to limit the impact of incorrect estimations.

Multiple flight tests are systematically conducted to verify the safety of vertical landing maneuver, long-range detection and robust tracking capability, and landing accuracy. The UAV lands vertically independent of ship motions in the same manner as a Navy helicopter lands on a ship. More than 100 landing tests are successfully conducted on a moving deck, which mimicked realistic and challenging 6 DOF ship motions. The machine learning object detector trained by the YOLOv3 algorithm begins identifying the 1.8 x 1.8 meters (6 x 6 feet) ship platform from 0.25 kilometers away, which means the range in the case of a real ship landing can be 17.3 kilometers (9.3 nautical miles). This is estimated assuming that the rear-side of a typical small ship occupies an area of 15 x 15 meters (50 x 50 feet). The unique nonlinear control system demonstrates robust tracking capability during a wide range of realistic scenarios such as random initial positions, complicated ship motions, communication latency, sensor noise, and in the presence of winds up to 9 m/s (17.5 knots). Since the landing command is triggered once the UAV flies into the pre-defined landing threshold area, the final landing points are randomly positioned within the area. The current landing threshold is 0.35 x 0.35 meters area which is sufficient to guarantee safe landing on a 1.22 x 1.22 meters (4 x 4 feet) landing deck.

Some of the key conclusions from this study are enumerated below.
\vspace{-1mm}
\begin{itemize}
  \setlength\itemsep{0em}
  \item The proposed vertical landing maneuver for VTOL UAVs, which involves not following the ship deck motions, was verified as a safe way of landing. This was achieved through multiple landing tests on the deck mimicking challenging ship motions (Oliver Hazard Perry frigate at sea state 6) and NATOPS helicopter ship landing operational limits (roll: $\pm$8, pitch: $\pm3$).
  \vspace{-1mm}
  \item The long-range vision system developed based on the state-of-the-art machine learning algorithm YOLOv3 demonstrated a 10 times greater detection range than the classical computer vision systems. The control system successfully utilized the detected object position and relative size as states for long-range tracking.
  \vspace{-1mm}
  \item The biggest challenge to implement the machine learning based object detection on the real-time autonomous flight was the time delay. To cope with the time delay issue, a long-range controller was constructed that responded less sensitively to errors around the setpoint and aggressively to large errors, using an exponential variation of feedback gain with error. This approach enabled the UAV to stay in the appropriate flight course while approaching the ship platform.
  \vspace{-1mm}
  \item The developed close-range vision system that combined the classical computer vision techniques and screening algorithms guaranteed fast and reliable detection of the visual cue and demonstrated precise relative position and orientation estimation. The update rate was approximately 15 times faster than the machine learning vision system and therefore, the control system was able to control the UAV more precisely using faster feedback.
  \vspace{-1mm}
  \item Even after going through the configured Kalman estimator, large/false estimation error can still occur from time to time. To prevent responding to such non-physical estimations, the probabilistic nonlinear controller was developed. It probabilistically perceives if the estimation is physically possible or not, based on the normal distribution curve and known UAV characteristics. Multiplying the estimation value by its probability can effectively reject responding to false estimations. By this approach, the controller never generated abrupt large control inputs even when the vision system provided inaccurate estimations. This greatly improved the robustness of tracking.
  \vspace{-1mm}
  \item The landing accuracy depends on the pre-defined landing threshold. The accuracy can be increased by setting a smaller threshold; however, this will take more time and requires more control effort. Considering the UAV and landing pad size, the appropriate landing threshold of 0.35 x 0.35 meters was set and the UAV successfully landed within this area every time during the 100+ flight experiments. 
\end{itemize}

The results demonstrate conclusively the feasibility of this novel autonomous approach and landing strategy for VTOL UAVs, which is inspired by the Navy helicopter ship landing. This is a significant accomplishment since there are no known efforts in the literature, which focused on automating the real helicopter ship landing procedure. The next goal of the study is to improve the wind disturbance rejection capability by combining the current control system with reinforcement learning techniques so that the aircraft could robustly fly through the ship wake during the approach and landing phases.

\section{Acknowledgments}

This work has been funded by the Center for Unmanned Aircraft Systems (C-UAS), a National Science Foundation Industry/University Cooperative Research Center (I/UCRC) under NSF award Numbers IIP-1161036 and CNS-1946890, along with significant contributions from C-UAS industry members.

\listoffigures
\listoftables


\begin{thebibliography}{9}


\bibitem{sanchez2014approach}Sanchez-Lopez, J., Pestana, J., Saripalli, S. \& Campoy, P. "An approach toward visual autonomous ship board landing of a VTOL UAV," {\em Journal Of Intelligent \& Robotic Systems}. \textbf{74}, 113-127 (2014)

\bibitem{xu2009research}Xu, G., Zhang, Y., Ji, S., Cheng, Y. \& Tian, Y. "Research on computer vision-based for UAV autonomous landing on a ship," {\em Pattern Recognition Letters}. \textbf{30}, 600-605 (2009)

\bibitem{truong2016vision}Truong, Q., Rakotomamonjy, T., Taghizad, A. \& Biannic, J. "Vision-based control for helicopter ship landing with handling qualities constraints," {\em IFAC-PapersOnLine}. \textbf{49}, 118-123 (2016)

\bibitem{holmes2016autonomous}Holmes, W. \& Langelaan, J. "Autonomous ship-board landing using monocular vision," {\em Proc. 72nd Am. Helicopter Soc Forum}. \textbf{2} pp. 36 (2016)

\bibitem{takahashi2017autonomous}Takahashi, Marc D and Whalley, Matthew S and Mansur, Hossein and Ott, Ltc Carl and Minor, MAJ Joseph S and Morford, MAJ Zachariah G M. "Autonomous Rotorcraft Flight Control with Multilevel Pilot Interaction in Hover and Forward Flight," {\em Journal of the American Helicopter Society}. \textbf{63}, 1--13 (2017)

\bibitem{yakimenko2002unmanned}Yakimenko, O., Kaminer, I., Lentz, W. \& Ghyzel, P. "Unmanned aircraft navigation for shipboard landing using infrared vision," {\em IEEE Transactions On Aerospace And Electronic Systems}. \textbf{38}, 1181-1200 (2002)

\bibitem{meng2019visual}Meng, Y., Wang, W., Han, H. \& Ban, J. "A visual/inertial integrated landing guidance method for UAV landing on the ship," {\em Aerospace Science And Technology}. \textbf{85} pp. 474-480 (2019)


\bibitem{lee2003simulation}Lee, Dooyong and Horn, Joseph and Sezer-Uzol, Nilay \& Long, Lyle. "Simulation of pilot control activity during helicopter shipboard operations,"{\em AIAA Atmospheric Flight Mechanics Conference and Exhibit}. (2003)

\bibitem{lee2005simulation}Lee, Dooyong and Horn, Joseph and Sezer-Uzol, Nilay \& Long, Lyle. "Simulation of helicopter shipboard launch and recovery with time-accurate airwakes,"{\em Journal of Aircraft}. (2005)


\bibitem{colwell2002maritime}Colwell, J. "Maritime helicopter ship motion criteria-Challenges for operational guidance," {\em Challenges For Operational Guidance-NATO RTO Systems Concepts And Integration Panel SCI-120. Berlin, Germany}. (2002)

\bibitem{tuttle1976study}Tuttle, Roy M. "A study of helicopter landing behavior on small ships," {\em Journal of the American Helicopter Society}. \textbf{2}, 2--11 (1976)

\bibitem{soneson2016simulation}Soneson, Gregory L and Horn, Joseph F and Zheng, Albert. "Simulation testing of advanced response types for ship-based rotorcraft," {\em Journal of the American Helicopter Society}. \textbf{61}, 1--13 (2016)

\bibitem{lumsden1998challenges}Lumsden, B., Wilkinson, C. \& Padfield, G. "Challenges at the helicopter-ship dynamic interface,"  (1998)

\bibitem{minotra2020analysis}Minotra, Dev and Feigh, Karen M. "An Analysis of Cognitive Demands in Ship-Based Helicopter-Landing Maneuvers," {\em Journal of the American Helicopter Society}. \textbf{74}, 1--11 (2020)

\bibitem{stingl1970vtol}Stingl, A. "Vtol aircraft flight system," (Google Patents,1970), US Patent 3,487,553

\bibitem{nato} NATO standard, "Helicopter operations from ships other than aircraft carriers(HOSTAC)," volume=I, 2017



\bibitem{girshick2014rich}Girshick, R., Donahue, J., Darrell, T. \& Malik, J. "Rich feature hierarchies for accurate object detection and semantic segmentation," {\em Proceedings Of The IEEE Conference On Computer Vision And Pattern Recognition}. pp. 580-587 (2014)

\bibitem{girshick2015fast}Girshick, R. "Fast r-cnn," {\em Proceedings Of The IEEE International Conference On Computer Vision}. pp. 1440-1448 (2015)

\bibitem{ren2016faster}Ren, S., He, K., Girshick, R. \& Sun, J. "Faster R-CNN: towards real-time object detection with region proposal networks," {\em IEEE Transactions On Pattern Analysis And Machine Intelligence}. \textbf{39}, 1137-1149 (2016)

\bibitem{liu2016ssd}Liu, W., Anguelov, D., Erhan, D., Szegedy, C., Reed, S., Fu, C. \& Berg, A. "Ssd: Single shot multibox detector," {\em European Conference On Computer Vision}. pp. 21-37 (2016)

\bibitem{redmon2016you}Redmon, J., Divvala, S., Girshick, R. \& Farhadi, A. "You only look once: Unified, real-time object detection," {\em Proceedings Of The IEEE Conference On Computer Vision And Pattern Recognition}. pp. 779-788 (2016)

\bibitem{redmon2017yolo9000}Redmon, J. \& Farhadi, A. "YOLO9000: better, faster, stronger.," {\em Proceedings Of The IEEE Conference On Computer Vision And Pattern Recognition}. pp. 7263-7271 (2017)

\bibitem{redmon2018yolov3}Redmon, J. \& Farhadi, A. "Yolov3: An incremental improvement," {\em ArXiv Preprint ArXiv:1804.02767}. (2018)

\bibitem{benjdira2019car}Benjdira, B., Khursheed, T., Koubaa, A., Ammar, A. \& Ouni, K. "Car detection using unmanned aerial vehicles: Comparison between faster r-cnn and yolov3," {\em 2019 1st International Conference On Unmanned Vehicle Systems-Oman (UVS)}. pp. 1-6 (2019)

\bibitem{wang2007vision}Wang, X., Pan, S., Song, Z. \& Shen, W. "Vision based analytic 3D measurement algorithm for the autonomous landing of unmanned helicopter on ship deck," {\em Optical Technique}. \textbf{33} pp. 264-267 (2007)

\bibitem{zishan2007computer}Zishan, W. \& Weiqun, S. "Computer vision scheme for autonomous landing of unmanned helicopter on ship deck [J]," {\em Journal Of Beijing University Of Aeronautics And Astronautics}. \textbf{6} (2007)

\bibitem{sharp2001vision}Sharp, C., Shakernia, O. \& Sastry, S. "A vision system for landing an unmanned aerial vehicle," {\em Proceedings 2001 ICRA. IEEE International Conference On Robotics And Automation (Cat. No. 01CH37164)}. \textbf{2} pp. 1720-1727 (2001)

\bibitem{jihong2004method}Jihong, L. "A Method for Estimating Position and Orientation of an Unmanned Helicopter Based on Vanishing Line Information [J]," {\em Computer Engineering And Applications}. \textbf{9} (2004)

\bibitem{lange2009vision}Lange, S., Sunderhauf, N. \& Protzel, P. "A vision based onboard approach for landing and position control of an autonomous multirotor UAV in GPS-denied environments," {\em 2009 International Conference On Advanced Robotics}. pp. 1-6 (2009)

\bibitem{lee2020vision}Lee, B., Saj, V., Benedict, M. \& Kalathil, D. "A Vision-Based Control Method for Autonomous Landing of Vertical Flight Aircraft On a Moving Platform Without Using GPS," {\em ArXiv Preprint ArXiv:2008.05699}. (2020)

\bibitem{daly2015coordinated}Daly, J., Ma, Y. \& Waslander, S. "Coordinated landing of a quadrotor on a skid-steered ground vehicle in the presence of time delays," {\em Autonomous Robots}. \textbf{38}, 179-191 (2015)

\bibitem{araar2017vision}Araar, O., Aouf, N. \& Vitanov, I. "Vision based autonomous landing of multirotor UAV on moving platform," {\em Journal Of Intelligent \& Robotic Systems}. \textbf{85}, 369-384 (2017)

\bibitem{lee2020development}Lee, B. \& Benedict, M. "Development and Validation of a Comprehensive Helicopter Flight Dynamics Code," {\em AIAA Scitech 2020 Forum}. pp. 1644 (2020)

\bibitem{lee2018helicopter}Lee, B. "Helicopter Autonomous Ship Landing System," (Texas A\&M University,2018)

\bibitem{ghamry2016real}Ghamry, K., Dong, Y., Kamel, M. \& Zhang, Y. "Real-time autonomous take-off, tracking and landing of UAV on a moving UGV platform," {\em 2016 24th Mediterranean Conference On Control And Automation (MED)}. pp. 1236-1241 (2016)

\bibitem{xia2020adaptive}Xia, K., Lee, S. \& Son, H. "Adaptive control for multi-rotor UAVs autonomous ship landing with mission planning," {\em Aerospace Science And Technology}. \textbf{96} pp. 105549 (2020)

\bibitem{vlantis2015quadrotor}Vlantis, P., Marantos, P., Bechlioulis, C. \& Kyriakopoulos, K. "Quadrotor landing on an inclined platform of a moving ground vehicle," {\em 2015 IEEE International Conference On Robotics And Automation (ICRA)}. pp. 2202-2207 (2015)

\bibitem{hu2015fast}Hu, B., Lu, L. \& Mishra, S. "Fast, safe and precise landing of a quadrotor on an oscillating platform," {\em 2015 American Control Conference (ACC)}. pp. 3836-3841 (2015)

\bibitem{kim2016landing}Kim, J., Jung, Y., Lee, D. \& Shim, D. "Landing control on a mobile platform for multi-copters using an omnidirectional image sensor," {\em Journal Of Intelligent \& Robotic Systems}. \textbf{84}, 529-541 (2016)

\bibitem{rodriguez2019deep}Rodriguez-Ramos, A., Sampedro, C., Bavle, H., De La Puente, P. \& Campoy, P. "A deep reinforcement learning strategy for UAV autonomous landing on a moving platform," {\em Journal Of Intelligent \& Robotic Systems}. \textbf{93}, 351-366 (2019)

\bibitem{lawther1988motion}Lawther, A. \& Griffin, M. "Motion sickness and motion characteristics of vessels at sea," {\em Ergonomics}. \textbf{31}, 1373-1394 (1988)

\bibitem{natops} Naval Air Systems Command, "Helicopter operating procedures for air-capable ships NATOPS manual," 2003

\bibitem{doerry2008ship}Doerry, A. "Ship dynamics for maritime ISAR imaging," (Sandia National Laboratories,2008)

\bibitem{est1}Zhang, Z. "Flexible camera calibration by viewing a plane from unknown orientations," {\em Proceedings Of The Seventh Ieee International Conference On Computer Vision}. \textbf{1} pp. 666-673 (1999)

\bibitem{est2}Araki, N., Sato, T., Konishi, Y. \& Ishigaki, H. "Vehicle’s orientation measurement method by single-camera image using known-shaped planar object," {\em Int. J. Innov. Comput. Inf. Control}. \textbf{7}, 4477-4486 (2011)


\bibitem{Couprie2003}Couprie, M. \& Bertrand, G. "Topological gray-scale watershed transformation," {\em Vision Geometry VI}. \textbf{3168} pp. 136-146 (1997)


\bibitem{forstner1987fast}Förstner, W. \& Gülch, E. "A fast operator for detection and precise location of distinct points, corners and centres of circular features," {\em Proc. ISPRS Intercommission Conference On Fast Processing Of Photogrammetric Data}. pp. 281-305 (1987)

\bibitem{lev1}Levenberg, K. "A method for the solution of certain non-linear problems in least squares," {\em Quarterly Of Applied Mathematics}. \textbf{2}, 164-168 (1944)

\bibitem{lev2}Marquardt, D. "An algorithm for least-squares estimation of nonlinear parameters," {\em Journal Of The Society For Industrial And Applied Mathematics}. \textbf{11}, 431-441 (1963)

\bibitem{video}Lee, B. "Intelligent Ship Landing," (https://youtu.be/ExkyUOdgYaw)

\bibitem{davis2009beaches}Davis Jr, R. \& FitzGerald, D. "Beaches and coasts," (John Wiley \& Sons,2009)

\bibitem{parrot}Parrot "Parrot ANAFI Quadrotor UAV," (https://www.parrot.com/us/drones/anafi)


\end{thebibliography}
\end{document}